\pdfoutput=1
\documentclass[sigconf]{aamas} 

\usepackage{amssymb}

\usepackage{balance} % for balancing columns on the final page
\usepackage[linesnumbered,ruled,vlined]{algorithm2e}
\SetKwInput{KwInput}{Input} %set the Input
\SetKwInput{KwOutput}{Output} %set the Output
\SetKwComment{Comment}{/*}{*/} %set the Output

\usepackage{subcaption} %figures
\usepackage{multirow} %table_rows
\usepackage{float} %multicolumn_figures
\usepackage{mathtools}
\usepackage{markdown}

\usepackage{soul}
\usepackage{balance} % for balancing columns on the final page
\usepackage{enumitem}
\usepackage{subcaption}
\doi{FDIK3367}

\makeatletter
\gdef\@copyrightpermission{
  \begin{minipage}{0.2\columnwidth}
   \href{https://creativecommons.org/licenses/by/4.0/}{\includegraphics[width=0.90\textwidth]{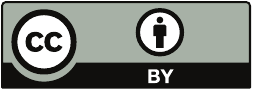}}
  \end{minipage}\hfill
  \begin{minipage}{0.8\columnwidth}
   \href{https://creativecommons.org/licenses/by/4.0/}{This work is licensed under a Creative Commons Attribution International 4.0 License.}
  \end{minipage}
  \vspace{5pt}
}
\makeatother

\setcopyright{ifaamas}
\acmConference[AAMAS '26]{Proc.\@ of the 25th International Conference
on Autonomous Agents and Multiagent Systems (AAMAS 2026)}{May 25 -- 29, 2026}
{Paphos, Cyprus}{C.~Amato, L.~Dennis, V.~Mascardi, J.~Thangarajah (eds.)}
\copyrightyear{2026}
\acmYear{2026}
\acmDOI{}
\acmPrice{}
\acmISBN{}

\acmSubmissionID{1757}

\title[AAMAS-2026 Formatting Instructions]{Issues with Measuring Task Complexity via Random Policies in Robotic Tasks}
\author{Reabetswe M. Nkhumise}
\affiliation{
  \institution{University of Sheffield}
  \city{Sheffield}
  \country{United Kingdom}}
\email{rabs.mike@yahoo.com}

\author{Mohamed S. Talamali}
\affiliation{
  \institution{University of Sheffield}
  \city{Sheffield}
  \country{United Kingdom}}
\email{m.s.talamali@sheffield.ac.uk}

\author{Aditya Gilra}
\affiliation{
  \institution{Wirtschaftsuniversit\"at}
  \city{Vienna}
  \country{Austria}}
\email{aditya.gilra@wu.ac.at}

\begin{abstract}
Reinforcement learning (RL) has enabled major advances in fields such as robotics and natural language processing. A key challenge in RL is measuring task complexity, which is essential for creating meaningful benchmarks and designing effective curricula. While there are numerous well-established metrics for assessing task complexity in tabular settings, relatively few exist in non-tabular domains. These include (i) Statistical analysis of the performance of random policies via Random Weight Guessing (RWG), and (ii) information-theoretic metrics Policy Information Capacity (PIC) and Policy-Optimal Information Capacity (POIC), which are reliant on RWG. In this paper, we evaluate these methods using progressively difficult robotic manipulation setups, with known relative complexity, with both dense and sparse reward formulations. Our empirical results reveal that measuring complexity is still nuanced. Specifically, under the same reward formulation, PIC suggests that a two-link robotic arm setup is easier than a single-link setup --- which contradicts the robotic control and empirical {RL} perspective whereby the two-link setup is inherently more complex. Likewise, for the same setup, POIC estimates that tasks with sparse rewards are easier than those with dense rewards. Thus, we show that both PIC and POIC contradict typical understanding and empirical results from {RL}. These findings highlight the need to move beyond RWG-based metrics towards better metrics that can more reliably capture task complexity in non-tabular RL with our task framework as a starting point. 
\end{abstract}


\newcommand{\BibTeX}{\rm B\kern-.05em{\sc i\kern-.025em b}\kern-.08em\TeX}

\begin{document}

%%% The following commands remove the headers in your paper. For final 
%%% papers, these will be inserted during the pagination process.

\pagestyle{fancy}
\fancyhead{}

%%% The next command prints the information defined in the preamble.

\maketitle 

%%%%%%%%%%%%%%%%%%%%%%%%%%%%%%%%%%%%%%%%%%%%%%%%%%%%%%%%%%%%%%%%%%%%%%%%
\section{Introduction}

Reinforcement Learning (RL) is a framework formulated to tackle sequential decision-making problems, which are at the heart of autonomous agents~\citep{Obando24}. It has demonstrated remarkable success in constructing decision-making agents across various domains, including games~\citep{Silver17,Jaderberg19}, robotics~\citep{Thomas18,Ibarz21} and energy systems~\citep{Yang20,Rocchetta19}, amongst others. Novel RL algorithms are continually being developed to tackle increasingly complex real-world problems.

% RL benchmarks & Curriculum learning
To evaluate progress, researchers employ a variety of RL \textit{benchmarks}~\citep{Bellemare13,Tassa18,Osband19} --- collections of tasks designed with varying levels of difficulty --- to assess and compare algorithms~\citep{Oller20}. When proposing a new method, it is standard practice to use such benchmarks to characterise its capabilities relative to existing approaches by measuring performance across tasks and examining how results scale with increasing difficulty. In addition, researchers also often leverage \textit{curriculum learning}, where agents are trained on a sequence of progressively more difficult tasks to enable intermediate learning and gradual skills acquisition~\citep{Narvekar21}. This approach has been shown to improve both learning and generalisability~\citep{Justesen18,Cobbe19,Rajan23}, with agents trained using curricula typically outperforming those trained directly on the most difficult tasks~\citep{Bengio09,Justesen18,Narvekar21}.   

% Curriculum learning
Measuring task difficulty is essential in both benchmarks and curriculum learning. In benchmarks, it ensures that tasks span a broad spectrum of challenges, providing comprehensive coverage that avoids sets that are uniformly trivial or overly difficult and thereby enables a more rigorous evaluation of algorithms’ capabilities~\citep{Duan16,Rajan23}. In curriculum learning, by contrast, task difficulty supports the structured ranking of tasks, allowing agents to encounter them in a progression that reflects their true hardness~\citep{Narvekar20}.

% Reinforcement learning (RL) 
{RL} tasks are commonly divided into tabular and non-tabular settings. Tabular RL assumes small, finite state-action spaces that can be explicitly enumerated, as in grid-world or bandit problems. Non-tabular RL, by contrast, involves large or continuous state-action spaces, typical of robotics control, autonomous driving, or energy management~\citep{Kormushev13,Dulac20,Rocchetta19,Furuta21}. 
% While measures of task difficulty exist for tabular settings, corresponding theory for non-tabular tasks remains limited~\citep{Conserva22}.
{While there are well-established metrics of task difficulty in tabular settings~\citep{Abel21,Conserva22}, a unified task complexity framework for non-tabular tasks is lacking~\citep{Conserva22}, largely since existing approaches rely on heuristics, make restrictive assumptions, or are computationally intractable~\citep{Conserva22,Conserva25}}. 

{Nevertheless, two notable approaches have been proposed to broadly analyse task complexity in non-tabular domains.}
% Recently, two notable approaches have been proposed to analyse the complexity of non-tabular RL tasks. 
Both are based on the Random Weight Guessing (RWG)~\citep{Schmidhuber99} process, in which untrained policies --- initialised with random weights --- are executed within tasks and their cumulative rewards (returns) are measured~\citep{Oller20}. The first approach employs \textit{statistical analysis} of the resulting return distributions to assess task difficulty~\citep{Oller20}. The second approach adopts an \textit{information-theoretic} perspective~\citep{Murphy12}, introducing two metrics: \textit{Policy Information Capacity} (PIC) and \textit{Policy-Optimal Information Capacity} (POIC)~\citep{Furuta21}. 

PIC quantifies the mutual information between the policy weights (parameters) and the returns. In contrast, POIC measures the mutual information between the policy parameters and an optimality variable, which indicates whether the agent behaves optimally throughout the episode~\citep{Furuta21}. A higher PIC value reflects a stronger dependence of returns on the random policy parameters, suggesting that the policy exerts a greater influence on performance and that the task is therefore easier. Similarly, high POIC values indicate that finding an optimal policy is relatively straightforward. Conversely, lower PIC and POIC values are associated with more difficult tasks.

% Knowns drawbacks of existing approaches
While the statistical approach provides a \textit{relative} measure of task complexity --- indicating, for instance, that one task is more difficult than another based on the return statistics of randomly sampled policies --- it does not quantify \textit{how {much} harder} one task is compared to another. This limitation is addressed by the information-theoretic approach through the PIC and POIC metrics, which offer quantitative measures of relative task complexity.

%Our paper reveals more limitations about PIC/POIC
In this work, we demonstrate that, despite the information-theoretic approach providing a more quantitative characterisation of relative task complexity, the resulting measures can be misleading in
{certain} 
% specific 
cases. 
Using tasks with \textbf{known relative complexity relationships}, we show that PIC and POIC can incorrectly capture task hardness. These tasks consist of simple robotic manipulation environments, \textit{1-link} and \textit{2-link} manipulators with one and two degrees of freedom (DoF) respectively --- where the objective is to control a robotic arm to reach specified target positions. Two reward formulations are considered: a \textit{dense} formulation, which provides incremental rewards as the arm approaches the target, and a \textit{sparse} formulation, which provides a non-negative reward only upon reaching the target. In our experiments, both PIC and POIC produced results that contradict expectations --- indicating, for instance, that the \textit{2-link} manipulator task is easier than the \textit{1-link} manipulator task under the same reward formulation, or that it is easier to find optimal policies in a sparse-reward setting than in a dense-reward one. These outcomes suggest that the PIC and POIC metrics may not be reliable indicators of task complexity, and that the question of reliable metrics remains open. We speculate that the inconsistencies of these metrics could be attributed to their reliance on the RWG process, which is known to be ineffective for tasks with sparse solution regions in the weight (parameter) space~\citep{Schmidhuber99}. Our contributions are threefold:
\begin{itemize}[leftmargin=*]
    \item We propose a framework for assessing task complexity metrics, achieved by using environments and reward formulations of known relative complexity. Specifically, we employ structurally comparable robotic manipulation environments evaluated under different reward formulations.
    \item Using this framework, we show that PIC and POIC yield results that contradict these known complexity relationships, suggesting that these metrics do not remain valid in certain task settings.
    \item We highlight the need for continued research into developing more reliable and interpretable measures of task complexity. 
\end{itemize}

The remainder of this paper is organised as follows. We briefly outline the methods used to assess task complexity --- i.e. statistical analysis of return distributions of random policies, and PIC and POIC --- in Section~\ref{tx:methodology}. Following this, we explicate the complexity of manipulation tasks in Section~\ref{tx:complex_manipulation}. Then, we present the results of task complexity analysis using the aforementioned methods on manipulation tasks in Section~\ref{tx:evaluation}. Finally, we discuss the limitations of these task complexity metrics in Section~\ref{tx:discussion}.   

%%%%%%%%%%%%%%%%%%%%%%%%%%%%%%%%%%%%%%%%%%%%%%%%%%%%%%%%%%%%%%%%%%%%%%%%
\section{Task Complexity Quantification Frameworks}
\label{tx:methodology}
In this section, we review the process used to determine the return statistics of random policies obtained through RWG, and the PIC and POIC task complexity metrics. Henceforth, we use the term \textit{return} interchangeably with \textit{performance}.  

\subsection{RWG and Statistical Analysis}
%overview of RWG and Performance distributions
The use of RWG, along with statistical analysis of performance, for analysing RL task complexity was introduced in~\cite{Oller20}. In this method, a policy model is represented by a neural network architecture. The model's parameters are randomly sampled at the beginning of each run and remain fixed thereafter. The untrained policy is then executed within the environment, and the resulting episodic rewards are recorded --- as outlined in Algorithm~\ref{algo:rwg_algorithm}. 
% \vspace{-1em}
\begin{algorithm}
\caption{Task Evaluation with RWG}\label{algo:rwg_algorithm}
    \KwInput{Prior distribution of parameters ${p(\theta)} = \mathcal{N}(0,I)$, Number of samples ${N}$, Number of episodes ${M}$.}
    \KwOutput{episodic cumulative reward ${S_{n,e}}$}
    {Initialize environment}\;
    {Create array \textit{$S_{n,e}$} of size ${N \times M}$}\;
	\For {$n = 1,2,\ldots,N$}
    {
        {Sample weights ${\theta_{n} \sim p(\theta)}$}\;
        \For {$e = 1,2,\ldots,M$}
        {
            {Reset the environment}\;
            {Run episode with ${\theta_{n}}$}\;
            {Store cumulative episode reward in ${S_{n,e}}$}\;
        }
    }
\end{algorithm}
% \vspace{-1em}

%review the maths
The prior distribution of parameters $p(\theta)$ is a multivariate normal distribution $\mathcal{N}(0,I)$, where $I \in \mathbb{R}^{d \times d}$ is an identity matrix over weight vectors $\theta_n \in \mathbb{R}^{d}$. $N$ is the number of parameter sets of the policy model, i.e. number of (random) policies. $M$ is the number of episodes per run. Performance of each policy indexed by $n$ is aggregated by computing the mean $M_{n}$ and variance $V_{n}$ of the cumulative rewards over its trial set of episodes, using~\citep{Oller20}: 
\begin{equation} \label{rwg_mean}
M_{n} = \frac{1}{M} \sum_{e=1}^{M} S_{n,e}
\end{equation}
\begin{equation} \label{rwg_var}
V_{n} = \frac{1}{M-1} \sum_{e=1}^{M}(S_{n,e} - M_{n})^{2} 
\end{equation}
% \begin{align} \label{rwg_stastic}
%     M_{n} &= \frac{1}{M} \sum_{e=1}^{M} S_{n,e} \\
%     V_{n} &= \frac{1}{M+1} \sum_{e=1}^{M}
% \end{align}
where ${S_{n,e}}$ is the episodic cumulative reward (i.e. performance sample) for the $e^{th}$ episode of the ${n^{th}}$ policy. For a given task environment, the aggregate performance of the policies is showcased in three plots:
\begin{enumerate}
    \item Log-scale histogram of $M_{n}$
    \item Mean performance $M_{n}$ vs rank $R_{n}$
    \item Performance variance ${V_{n}}$ vs mean performance $M_{n}$ 
\end{enumerate}
where rank $R_{n}$ sorts the policies according to performance, with 1 denoting the policy with the lowest mean performance and larger values representing policies with higher mean performances. If two policies rank the same, then the tie is broken by ranking them in the order in which their weights were sampled. 

\subsection{PIC and POIC}
%overview of PIC
\textit{Mutual information} is a quantity that measures the dependency between two random variables~\citep{Cover06,Murphy12}. Unlike the correlation coefficient, it is not limited to only linear relationships but can also describe nonlinear ones~\citep{Murphy12}. PIC is the mutual information between the policy model parameters and the corresponding episodic cumulative rewards (i.e. \textit{return} samples). It is given by~\citep{Furuta21},
\begin{equation} \label{PIC_def}
 \mathcal{I}(R; \Theta) = \mathcal{H}(R) - \mathbb{E}_{p(\theta)} [\mathcal{H}(R | \Theta = \theta)]
\end{equation}
where $\mathcal{H}(\cdot)$ is Shannon entropy, $R$ is the episodic cumulative reward random variable, and $\Theta$ is the random variable of policy model parameters. Intuitively, if the parameters $\Theta$ do not tightly determine $R$ (i.e. have little effect on the reward signal), then the first term and second term in Equation~\ref{PIC_def} will be approximately equal. That is, $\mathcal{H}(R) \approx  \mathbb{E}_{p(\theta)} [\mathcal{H}(R | \Theta = \theta)]$ and hence ${\text{PIC}  \approx 0}$. In that case, the task is relatively hard and therefore, ${\text{PIC} \to 0}$ as tasks become harder. 

%overview of POIC
POIC is the mutual information between the policy model parameters and the  \textit{optimality variable}. An optimality variable represents whether the agent behaves optimally during the entire episode~\citep{Furuta21}. For instance, when POIC is expressed as~\citep{Furuta21}, 
\begin{equation} \label{POIC_def}
 \mathcal{I}(\mathbb{O}; \Theta) = \mathcal{H}(\mathbb{O}) - \mathbb{E}_{p(\theta)} [\mathcal{H}(\mathbb{O} | \Theta = \theta)]
\end{equation}
$\mathbb{O}$ is the optimality variable which $\mathbb{O} = 1$ when the agent behaves optimally during the episode, and $\mathbb{O} = 0$ otherwise. If parameters $\Theta$ have significant effect on the agent's optimal performance, then the difference between the first and second terms in Equation~\ref{POIC_def} will be large. This would highlight the ease of acting optimally in the task. 

%PIC and POIC
Note that PIC and POIC are aligned. They respectively represent the influence of the policy model parameters $\Theta$ on rewards and optimal behaviour. Furthermore, they can be viewed as the remaining randomness in rewards or optimality after accounting for the randomness caused by the parameters $\Theta$. The residual randomness reflects variability inherent to the environment. Both PIC and POIC make use of performance samples generated via Algorithm~\ref{algo:rwg_algorithm}. 
In practice, PIC and POIC are empirically estimated by discretising the return distribution $p(R)$ and each conditional distribution $p(R| \theta_{i})$ into $B$ identical bins, as follows --- starting with PIC~\citep{Furuta21}:
\begin{equation} \label{PIC_def_imperical}
\begin{aligned}
\hat{\mathcal{I}}(R; \theta) = &- \sum_{b=1}^{B} \hat{p}(R_{b}) \log(\hat{p}(R_{b})) \\
&+ \frac{1}{N} \sum_{n=1}^{N} \sum_{b=1}^{B} \hat{p}(R_{b} | \theta_{n}) \log(\hat{p}(R_{b} | \theta_{n}))
\end{aligned} 
\end{equation}
where $N$ is the number of random policies and $\hat{p}(R_{b})$ estimates a portion of return samples in bin $b$ with respect to the total number of return samples. For a given $\theta_{n}$, the fraction of return samples in bin $b$ relative to all return samples is $\hat{p}(R_{b} | \theta_{n})$. POIC is estimated using~\citep{Furuta21}:
\begin{equation}\label{POIC_def_imperical}
\begin{aligned}
\hat{\mathcal{I}}(O; \theta) &= -\hat{p}_{1} \log(\hat{p}_{1}) - (1-\hat{p}_{1}) \log(1-\hat{p}_{1}) \\
&+ \frac{1}{N} \sum_{n=1}^{N}  \left[ \hat{p}_{1n} \log(\hat{p}_{1n}) + (1-\hat{p}_{1n}) \log(1-\hat{p}_{1n})\right]
% \hat{I}(O; \theta) &= -\hat{p}_{1} \log(\hat{p}_{1}) - (1-\hat{p}_{1}) \log(1-\hat{p}_{1}) \\
% &+ \frac{1}{N}\left( \sum_{n=1}^{N} \hat{p}_{1n} \log(\hat{p}_{1n}) + (1-\hat{p}_{1n}) \log(1-\hat{p}_{1n})\right)
% p(\mathbb{O} = 1 | \theta_{n}) &\approx \frac{1}{M} \sum_{e=1}^{M} \exp\left( \frac{ S_{n,e} - S_{max}}{\lambda} \right) \doteq \hat{p}_{1n} \\
% p(\mathbb{O} = 1) &\approx \frac{1}{N} \sum_{n=1}^{N} \hat{p}_{1n} \doteq  \hat{p}_{1}
\end{aligned} 
\end{equation}
where $\hat{p}_{1} \doteq p(\mathbb{O} = 1) \approx \frac{1}{N} \sum_{n=1}^{N} \hat{p}_{1n}$ and $\hat{p}_{1n} \doteq  p(\mathbb{O} = 1 | \theta_{n}) \approx \frac{1}{M} \sum_{e=1}^{M} \exp\left( \frac{ S_{n,e} - S_{max}}{\lambda} \right)$. Note that $\lambda$ is a temperature parameter and $S_{max} = \max[ S_{n,e}, S^{*} ]$, where $S^{*}$ is the episodic cumulative reward of an optimal policy $\pi^{*}$. Minimum and maximum values in the return samples are set as limits and divided into $B$ equal parts for the calculations. 

We now consider the complexity of robotic reaching tasks that we will use to evaluate
the task complexity metrics presented in this section.
% these metrics.

%%%%%%%%%%%%%%%%%%%%%%%%%%%%%%%%%%%%%%%%%%%%%%%%%%%%%%%%%%%%%%%%%%%%%%%%

\section{Complexity of robotic reaching tasks}\label{tx:complex_manipulation}
\begin{figure}[h!]
\centering
\begin{tabular}{@{}c c@{}}
    \includegraphics[width=0.19\textwidth]{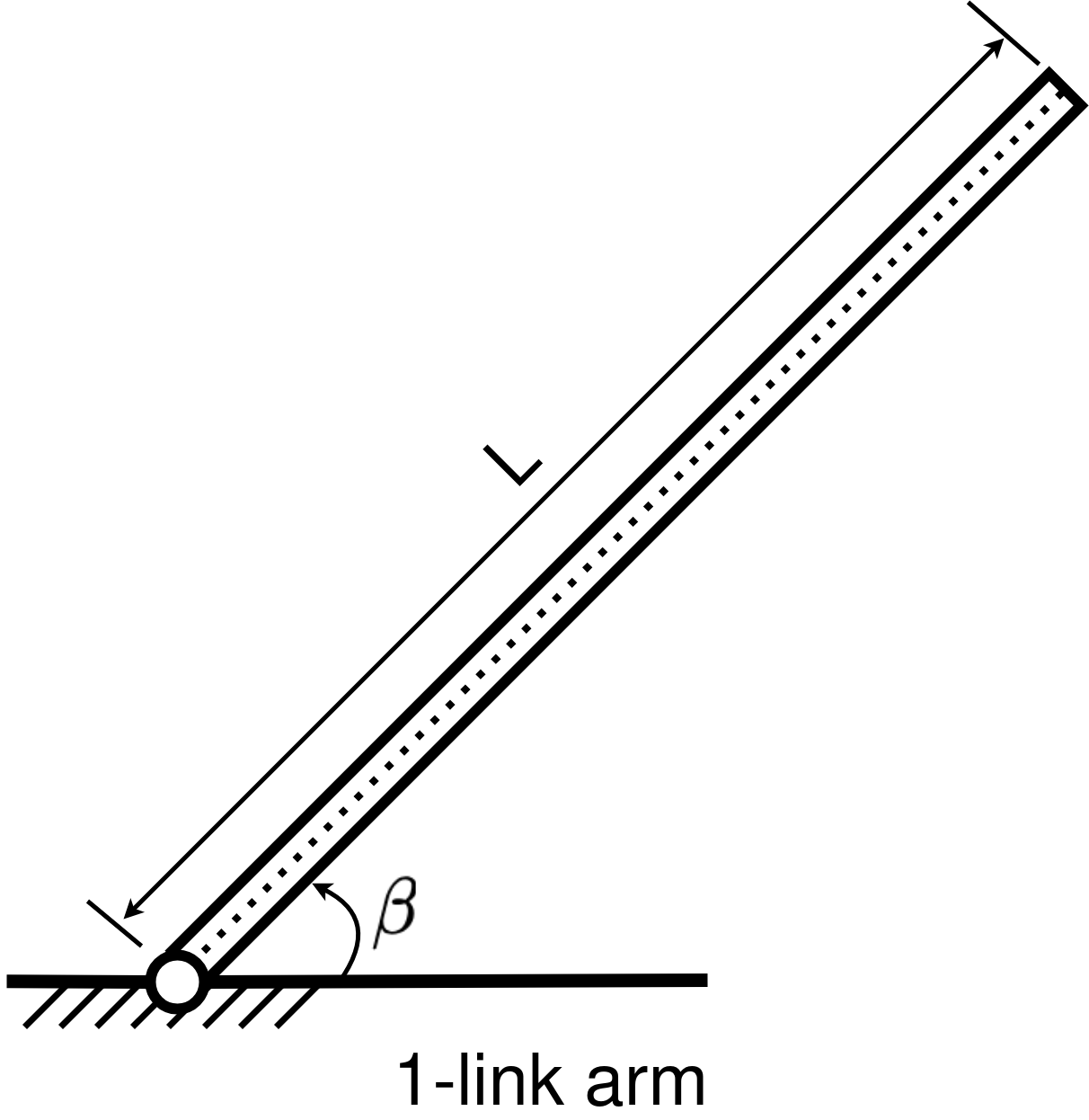} &
    \includegraphics[width=0.16\textwidth]{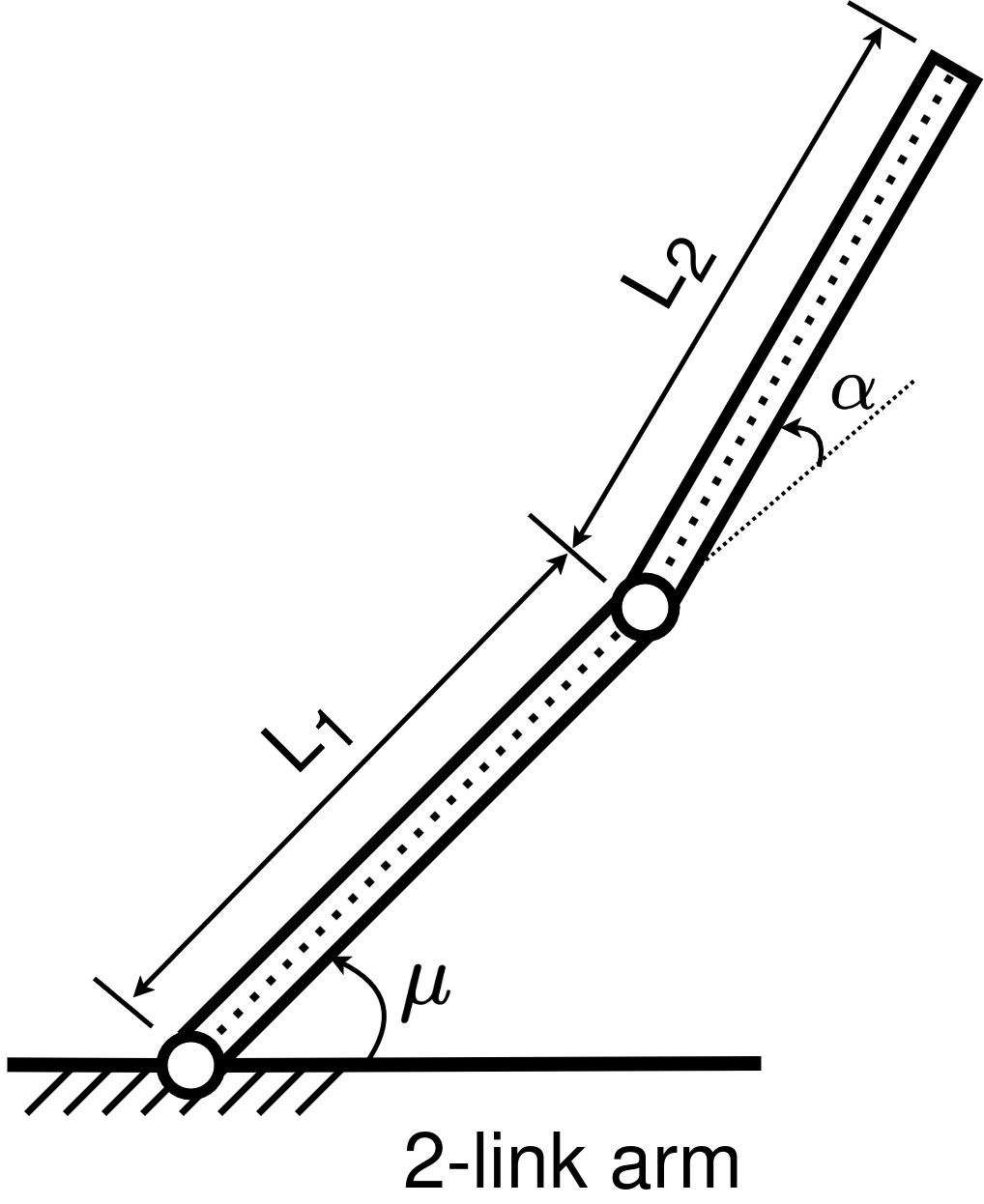} 
\end{tabular}
\vspace{-1em}
\caption{Illustration of manipulators.}\label{arm_illustration}
\vspace{-1em}
\end{figure}

%introduce manipulators
Robot manipulation is a classic problem that has been studied rigorously in control theory~\citep{Siciliano08,Mason18}. Within manipulation, reaching tasks are defined by the objective of moving the end-effector to desired positions.  
We consider such tasks along with fully actuated serial manipulators shown in Figure~\ref{arm_illustration}, where the motion of each joint is directly controllable. A manipulator with \textit{n} number of links or joints, often referred to as an \textit{n-link} arm, has the dynamic model~\citep{Corke11}:
\begin{equation}\label{manipulator_dynamics}
\begin{aligned}
    M(q)\ddot{q} + C(q,\dot{q})\dot{q} + G(q) + F(\dot{q}) + J(q)^{T}\eta = \tau
\end{aligned}    
\end{equation}
where $q, \dot{q}, \ddot{q} \in \mathbb{R}^{n}$ are joint position, velocity and acceleration vectors, respectively. $\tau \in \mathbb{R}^{n}$ is actuator joint torque vector, $M(q) \in \mathbb{R}^{n\times n}$ is inertia matrix, $C(q,\dot{q})\dot{q}  \in \mathbb{R}^{n}$ is Coriolis and centrifugal vector, $G(q) \in \mathbb{R}^{n}$ is gravity vector, and $F(\dot{q}) \in \mathbb{R}^{n}$ is friction torque. $J(q)^{T}$ is the transpose of the Jacobian matrix that relates $\eta \in \mathbb{R}^{6}$ forces at the end-effector to joint torques. 

%impact of DoF on complexity
Note that a fully actuated \textit{n-link} arm has $n$ degrees-of-freedom (DoF), i.e. \textit{n}-DoF. Higher \textit{n}-DoF enable agile, precise and energy-efficient robot motions~\citep{Murray17}, but consist of larger state-action space dimensionality. This leads to more complex dynamics~\citep{Copot18} that contribute to the task complexity~\citep{Hentout23}. For instance, as $n$ increases the system characteristics are impacted as follows:
\begin{enumerate}[wide, labelwidth=!, labelindent=0pt]
\item The number of coupled terms in $M(q)$, $C(q,\dot{q})$, $F(\dot{q})$ and $G(q)$ increases~\citep{Murray17}. This means the robot dynamics become highly nonlinear and more complex, resulting in unpredictable behaviour under perturbations~\citep{Spong06, Sciavicco12}.  
\item The coordination of multiple joints becomes more intricate and necessary to avoid factors such as link collisions, joint limits and singularity~\citep{Chiaverini02,Nakanishi08,Siciliano09}.  This means controlling the system becomes more difficult.
\item The ability of the robot to move in arbitrary directions, called manipulability, increases~\citep{Yoshikawa85,Vahrenkamp12,Khadem18}. This means the set of possible target positions for the end-effector grows. This comes with high computational control effort~\citep{Nakanishi08}, since the algorithms are burdened with learning more target positions.  
\end{enumerate}

%declaring complexity_ranking
Solving Equation~\ref{manipulator_dynamics} to compute joint motion requires algorithm complexity of $\mathcal{O}(n)$~\citep{Featherstone08}. This illustrates how the computational costs of the dynamics scale with \textit{n}-DoF.  In general, it can be declared that controlling an \textit{n+1-link} manipulator is inherently more difficult than controlling an \textit{n-link} manipulator. In summary,
\begin{equation}\label{task_difficulty_manipulator}
    \mathcal{C}(\text{\textit{n-link} manipulator}) < \mathcal{C}(\text{\textit{n+1-link} manipulator})   
\end{equation}
where $\mathcal{C}(\cdot)$ denotes the hardness of controlling the system. Although Equation~\ref{task_difficulty_manipulator} is not quantitative, 
it is useful for sanity check. In the next section, ranking of tasks via Equation~\ref{task_difficulty_manipulator} 
% based on complexity of manipulation,  
will be compared with those provided by the earlier task complexity metrics.
% it can usefully guide the ranking of tasks in hardness levels. In the next section, ranking of tasks via Equation~\ref{task_difficulty_manipulator} will be compared with those provided by PIC and POIC. 

%%%%%%%%%%%%%%%%%%%%%%%%%%%%%%%%%%%%%%%%%%%%%%%%%%%%%%%%%%%%%%%%%%%%%%%%

% Here is a statement.\footnote{\label{fn:myfoot}This is the footnote text.}

% Later I want to point to the same footnote\textsuperscript{\ref{fn:myfoot}}.
% \ref{}

% Code and Supplementary material:~\url{https://github.com/nkhumise-rea/task_complexity.git}

\section{Experimental evaluation}\label{tx:evaluation}
In this section, the methods described in Section~\ref{tx:methodology} are evaluated in a class of reaching tasks\footnote{\label{fn:code1}Code and Supplementary material are available at:~\url{https://github.com/nkhumise-rea/task_complexity.git}}. The purpose of the experiments is to answer the following questions:
(1) \textit{Can the statistical analysis of the performance of random policies and PIC/POIC effectively capture the task complexity of reaching tasks?}
(2) \textit{Do task difficulty levels ranked by PIC/POIC align with the ranking suggested by Equation~\ref{task_difficulty_manipulator}? } 

Section~\ref{experimetal_setup} describes the experimental setup, while Section~\ref{framework_tasks} introduces the task framework used for assessing the accuracy of the task complexity metrics. 
% presented in Section~\ref{tx:methodology}.
In Section~\ref{RL_Analysis}, we train RL agents on the tasks defined within this framework and use their performance to verify the tasks' complexity.
% to compute the corresponding complexity metrics. 
The resulting measures are then compared with known complexity rankings from robotics. Section~\ref{Task Complexity Analysis} examines the limitations of the task complexity metrics, showing that PIC and POIC can inaccurately measure task complexity.

\subsection{Experimental Setup}\label{experimetal_setup}
%description of the setup
We employ manipulators shown in Figure~\ref{arm_illustration} in six task settings (discussed in Section~\ref{framework_tasks}). For each task, both the end-effector and its target position are randomly initialised at the start of each episode. 
This is a good training practice that prevents environment overfitting when training RL agents~\citep{Whiteson11}. 
Friction is ignored, states are assumed to be fully observable, and the arms are operated on a horizontal plane; hence, the effects of gravity are 
% \textcolor{red}{ignored}.
not considered.

The arm configurations are evaluated on dense- and sparse-rewards. In dense-reward settings, the reward function is 
% $r = -\omega_1\lVert P_{ee} - P_{g} \rVert_{2}^{2} - \omega_2\lVert action \rVert_{2}^{2}$
\begin{equation}\label{reward_function_dense}
    r = -\omega_1\lVert P_{ee} - P_{g} \rVert_{2}^{2} - \omega_2\lVert action \rVert_{2}^{2}
\end{equation}
where $[\omega_1,\omega_2] = [1,1]$ are distance and control weights. $P_{ee}$ and $P_{g}$ are respectively end-effector and target/goal positions. In sparse-reward settings, 
% the reward function is
% \begin{equation}\label{reward_function_sparse}
%     r = \begin{dcases}
%         ~~~~0, &~\text{if } ~\lVert P_{ee} - P_{g} \rVert_{2} < 0.05 ~\text{meters} \\
%         -1, &~\text{otherwise}
%     \end{dcases}
% \end{equation}
% where $0.05$ meters is the threshold distance between the end-effector and target position. 
$r = 0$ when the end-effector is within the threshold distance from target ($< 0.05$ meters), otherwise $r= -1$. 
All the tasks have a maximum of $50$ steps per episode, $500$ training episodes and $N = 10^{4}$ samples (i.e. random policies). To match the experimental setup in~\citep{Furuta21}, the policy network consists of 2 hidden layers with $32$ neurons each and the number of discretisation bins $B = 10^{5}$.

\subsection{Tasks framework}\label{framework_tasks}
Our framework consists of six tasks that include three arm setups, each with dense- and sparse-rewards. The arm setups include: (1) \textit{1-link} arm with link length $L = 1.00$ meter, (2) \textit{1-link} arm with link length $L = 1.65$ meters, and (3) \textit{2-link} arm with link lengths $L_1 = 0.95$ and $L_2 = 0.70$ meters. We ensured that the link lengths of \textit{2-link} arm sum to $1.65$ m, to have the same total length as the \textit{1-link} arm in (2) above. This makes these two arms (2) and (3) have equivalent magnitude of error at the end-effector, leading to rewards that can be directly comparable, while varying in complexity only due to the number of links/joints. 

The error in the end-effector arises from the errors in the arm joint angles being amplified by the link lengths, which results in higher positional errors at the end-effector for longer link lengths, as captured by (for small errors):
\begin{equation}\label{error_bounds}
    \left\| \delta{x} \right\|_{2} = \left\| \sum_{k=1}^{n} J_{k}(\theta) \delta{\theta_k} \right\|_{2} \leq \epsilon \sum_{i=1}^{n} il_i
    % \left\| \delta{x} \right\|_{2} &= \left\| \sum_{k=1}^{n} J_{k} \delta{\theta_k} \right\|_{2}
\end{equation}
where $\delta{x}$ is error at the end-effector, $\delta{\theta}_{k}$ is angle error of the $k-$th joint, and $J_{k}(\theta)$ is the $k-$th column of the Jacobian matrix. $n$ is the number of DoF, while $l_i$ is the $i-$th link on the arm. $\epsilon$ is the worst-case joint error, i.e. $| \delta{\theta_k}| \leq \epsilon$ for $k=1,\cdots,n$ 
% (see Appendix~\ref{appx:ee_error} for details on Equation~\ref{error_bounds}). 
(see Supplementary material\textsuperscript{\ref{fn:code1}} for details on Equation~\ref{error_bounds}). 
Generally, tasks with higher error rates present greater learning challenges for RL algorithms, thus resulting in reduced rewards or longer learning time~\citep{Wang20,Edmondson25}.

% \textbf{Structural Similarity}
We selected the three arm configurations to study independently the effects of altering link lengths and number of joints. This simplifies task comparison and ensures task structural homogeneity. Following Equation~\ref{task_difficulty_manipulator}, we expect \textit{1-link} arm tasks to be easier than the \textit{2-link} arm task 
% --- irrespective of reward formulation. 
under the same reward formulation. 
From Equation~\ref{error_bounds}, we expect the \textit{1-link} arm task with a shorter link length to be easier than the \textit{1-link} arm task with a longer link length. Additionally, we expect consistency with RL literature~\citep{Andrychowicz17,Pathak17,Sutton18}, where tasks in dense-reward settings are easier than those in sparse-reward settings. 

\textbf{Importance of structurally similar tasks. }
Compared to most RL benchmarks, often the tasks are varied (as they should) but not structurally related~\citep{Duan16,Bellemare13}, e.g. Cartpole and MountainCar in OpenAI Gym~\citep{Brockman16}. As such, most benchmarks can be unsuitable for reliably assessing new task complexity methods. 
To ensure meaningful validation, methods should first be evaluated on families of closely related tasks with known relative complexity (such as our setup). 
This enables sanity checking, clearer interpretation of results, and easier characterisation of the proposed methods prior to applying them to large heterogeneous benchmarks. 

We start by comparing learning curves across our task framework for a baseline algorithm, to confirm if these match our expectations.

\subsection{Reinforcement Learning of Tasks}\label{RL_Analysis}
\begin{figure*}[th!]
\centering
\begin{tabular}{@{}l r@{}}
    \includegraphics[width=0.35\textwidth]{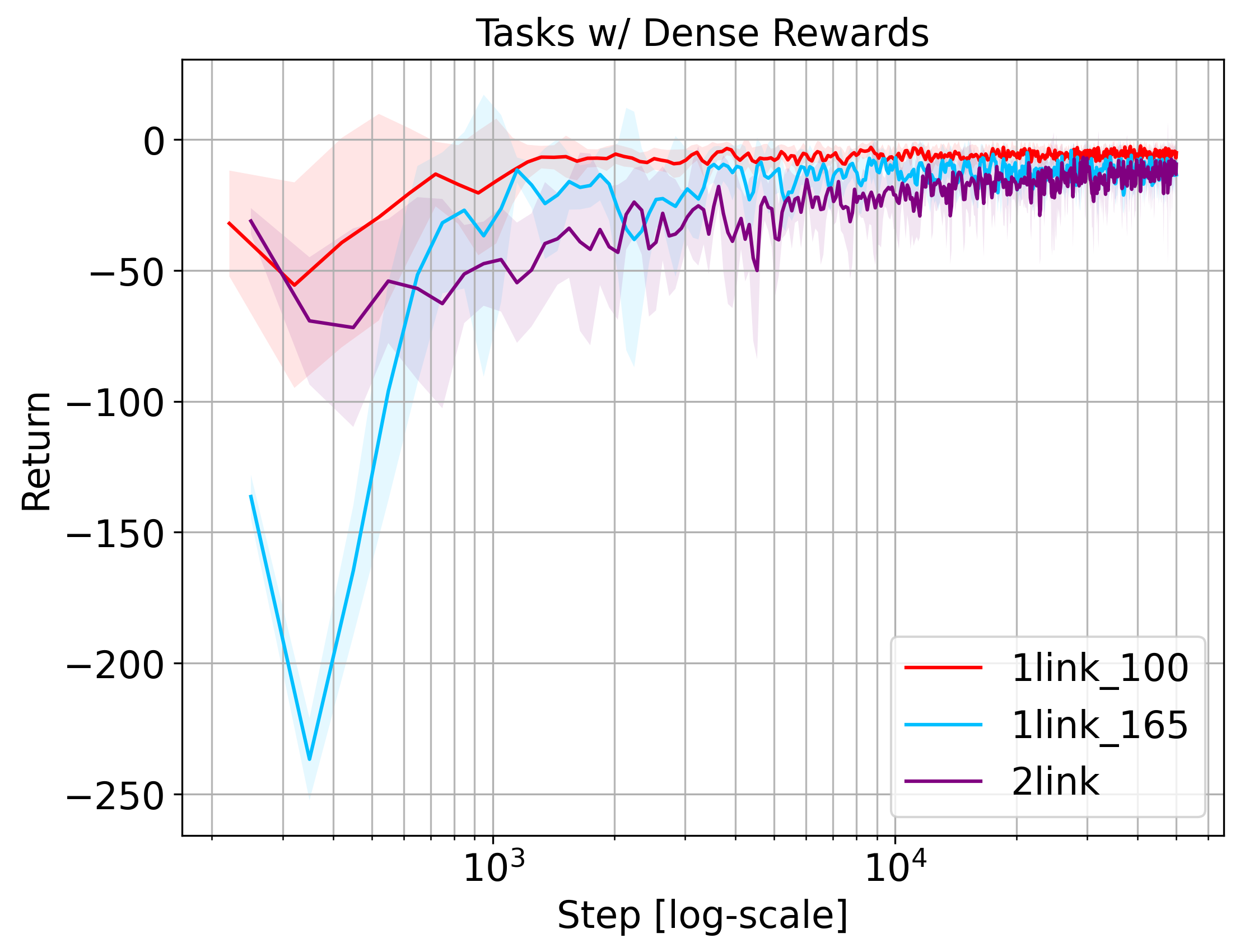} &
    \includegraphics[width=0.35\textwidth]{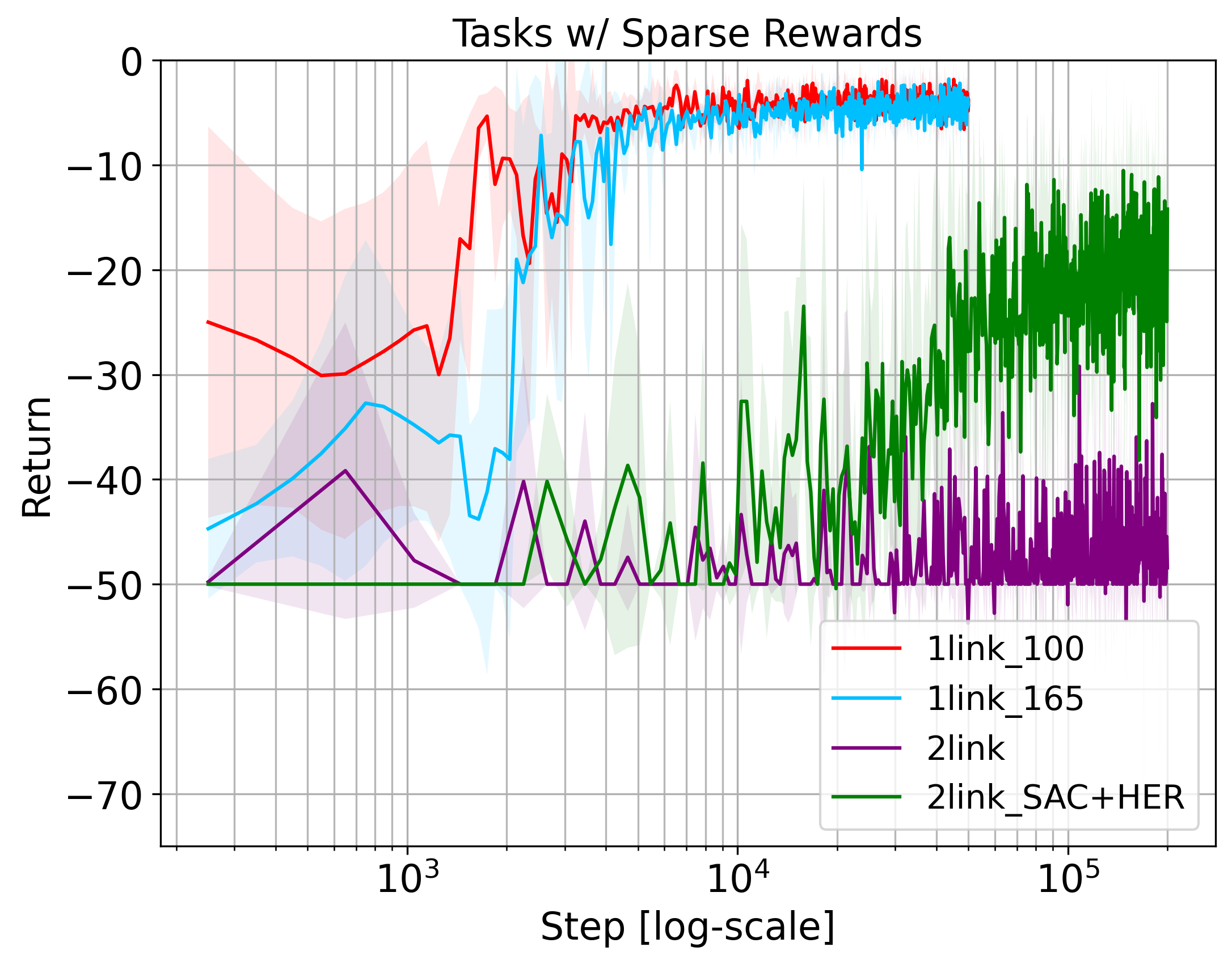} 
\end{tabular}
\vspace{-1em}
\caption{Learning curves of SAC algorithm across the six tasks. The left panel depicts agent performance in dense-reward settings, while the right panel is in sparse-reward settings. To accommodate wide and varying ranges of steps, results are plotted on a logarithmic scale to enhance interpretability. In the \textit{2-link} arm with sparse rewards, SAC results are presented with HER~\citep{Andrychowicz17} augmentation (SAC+HER) and without it. {The results are obtained via evaluation of each task over 5 runs.}}\label{RL_learning_across}
\end{figure*}

We compare learning curves of a state-of-the-art algorithm Soft Actor Critic (SAC)~\citep{Haarnoja18} across the tasks in our task framework. Studying how the algorithm performs during training provides us with information about convergence, such as the optimal return and convergence time for the tasks. The learning curves are presented in Figure~\ref{RL_learning_across}. 
Details about the SAC architecture and its hyperparameters are provided in the Supplementary material\textsuperscript{\ref{fn:code1}}. Note that the same network model for SAC was employed in all the tasks. 

\textbf{Dense-reward settings. }
We note in Figure~\ref{RL_learning_across} that \textit{1-link} arm with link length $L=1.0$ m converges faster and to a higher return than \textit{1-link} arm with link length $L=1.65$ m. Similarly, in the \textit{1-link} arm with link length $L=1.65$ m, the agent converges faster than in the \textit{2-link} arm task. This highlights the order of tasks based on their hardness (from easiest to hardest) to be \textit{1-link} arm ($L=1$ m), \textit{1-link} arm ($L=1.65$ m) and \textit{2-link} arm. This is consistent with our expectations as supported by Equations~\ref{task_difficulty_manipulator} and ~\ref{error_bounds}. 

\textbf{Sparse-reward settings. }
In Figure~\ref{RL_learning_across}, we see that pure SAC is unable to solve the \textit{2-link} arm task, showing that \textit{2-link} arm task is harder than the \textit{1-link} arm tasks, aligning with Equation~\ref{task_difficulty_manipulator}. Even SAC augmented with {Hindsight} Experience Replay (HER)~\citep{Andrychowicz17} takes significantly longer to converge in the \textit{2-link} arm task compared with \textit{1-link} arm tasks. Further, SAC converges faster in the \textit{1-link} arm ($L=1$ m) task than in \textit{1-link} arm ($L=1.65$ m) task. The ordering of these three arms is consistent with our expectations, in this sparse-reward settings as well.

\textbf{Dense-reward vs Sparse-reward settings. }
By comparing the dense- and sparse-reward settings, we observe that the algorithm converges quicker in dense-reward settings than in sparse-reward settings. Furthermore, SAC performance is noisier in the sparse-reward settings than its counterpart. This coincides with intuition that tasks with dense rewards are easier than with sparse rewards.

\vspace{0.5em}
REMARK 1.~\textit{In our settings, SAC(+HER) could solve the tasks (verified by demonstrations). However, in general, algorithms may fail to solve tasks or perform optimally. This restricts the usage of learning curves in assessing task complexity.}
\vspace{0.5em}

In the subsequent section, we compare methods introduced in Section~\ref{tx:methodology} for quantifying task complexity. These methods are independent of specific RL algorithms.  
\vspace{-2.0em}

\subsection{Task Complexity Analysis}\label{Task Complexity Analysis}
In this section, we categorise our results into examining how task complexity is influenced by a) link length, b) number of DoF, and c) reward formulation. We compare measures of task complexity beginning with statistical analysis of performance, and then following with PIC and POIC.   

\underline{\textsc{(I) Statistical analysis of performance. }} 
To carry out this analysis, we used Algorithm \ref{algo:rwg_algorithm} to capture the cumulative rewards (returns or performance) of $N=10^{4}$ randomly sampled policies via RWG~\citep{Oller20}. The performance was then aggregated into mean $M_{n}$ and variance $V_{n}$ using Equations \ref{rwg_mean} and~\ref{rwg_var}. We visualise the aggregated performance in three plots: \textit{mean performance histograms} (Log-scale histogram of $M_{n}$), \textit{mean performance curve} ($M_{n}$ vs $R_{n}$ plot), and \textit{variance distribution} ($\sqrt{V_{n}}$ vs $M_{n}$ plot). Note that $R_{n}$ is rank, from lowest to highest mean performance. 

\begin{figure*}[h!] 
\centering
     \raisebox{1.0cm}{\rotatebox{90}{\textbf{1-link arm [$1.0$] (dense)}}}
     \hspace{.125cm}
     \begin{subfigure}[b]{0.73\textwidth}
         \centering
         \includegraphics[width=\textwidth]{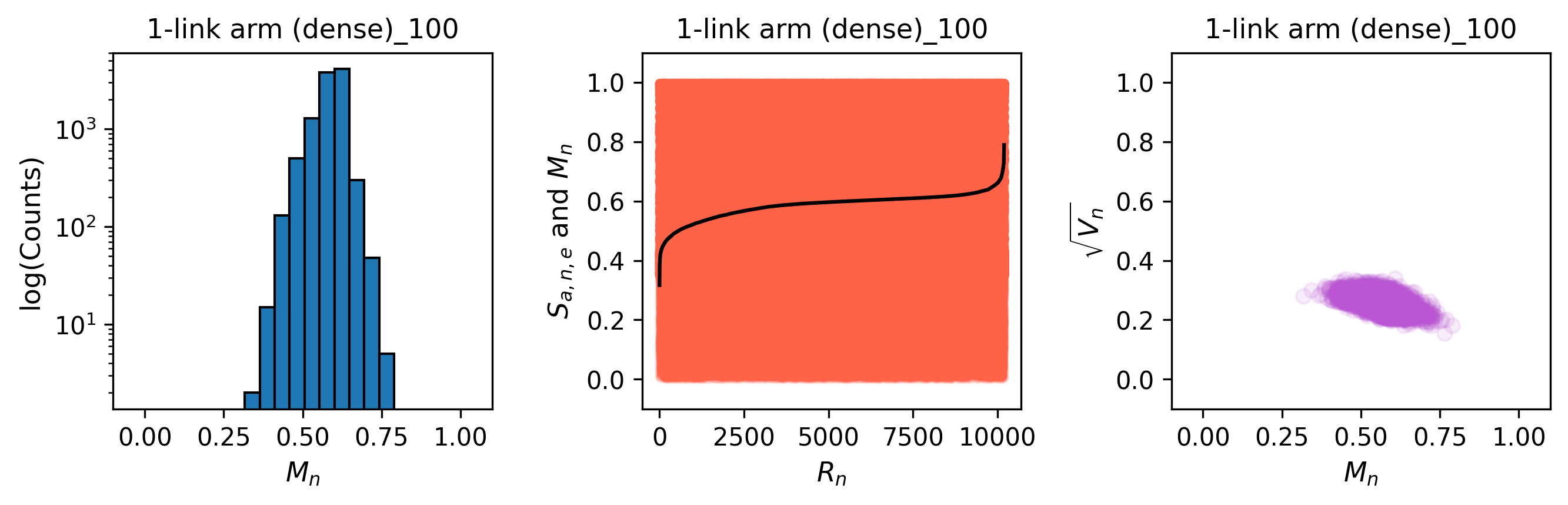} 
         \caption{}
     \end{subfigure}
     \par
     \raisebox{1.0cm}{\rotatebox{90}{\textbf{1-link arm [$1.65$] (dense)}}}
     \hspace{.125cm}
     \begin{subfigure}[b]{0.73\textwidth}
         \centering
         \includegraphics[width=\textwidth]{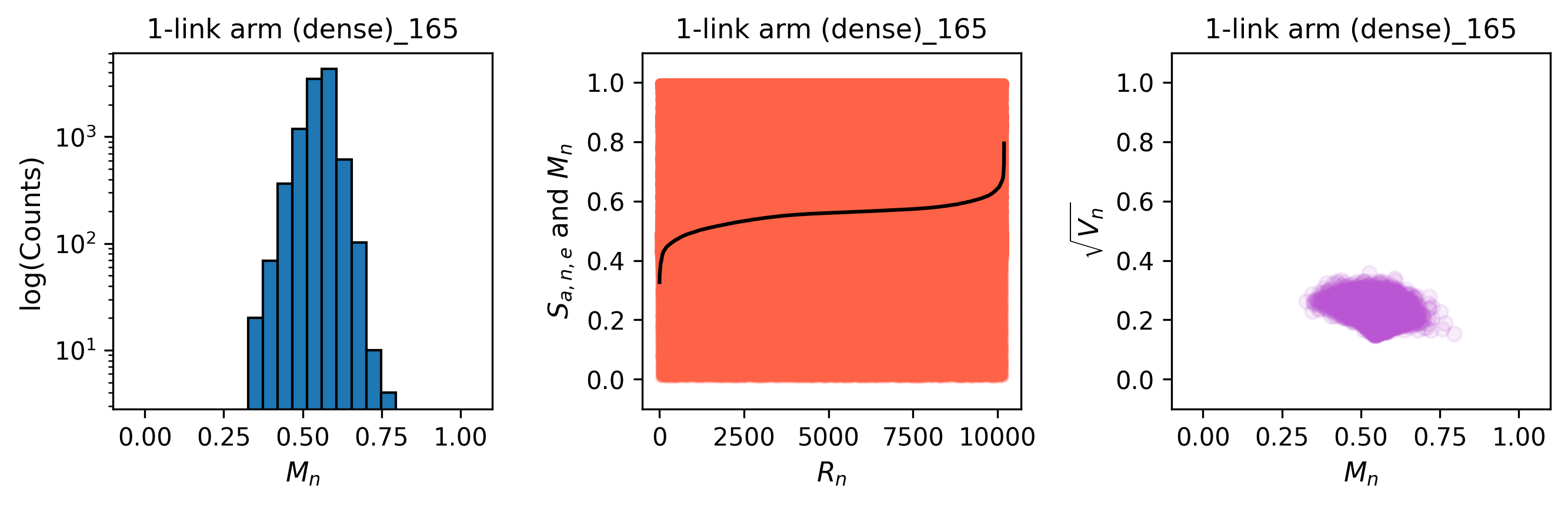}
         \caption{}
     \end{subfigure}
     \par
     \raisebox{1.25cm}{\rotatebox{90}{\textbf{2-link arm (dense)}}}
     \hspace{.125cm}
     \begin{subfigure}[b]{0.73\textwidth}
         \centering
         \includegraphics[width=\textwidth]{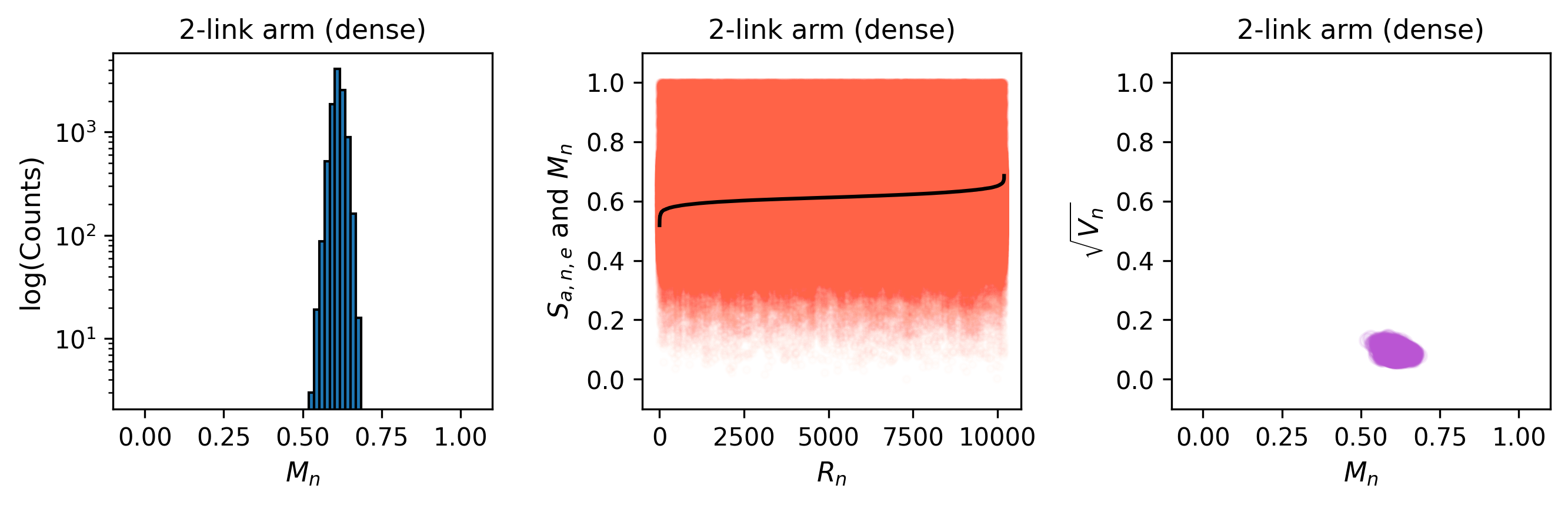}
         \caption{}
     \end{subfigure}
     \par
     \raisebox{1.25cm}{\rotatebox{90}{\textbf{2-link arm (sparse)}}} 
     \hspace{.125cm}
     \begin{subfigure}[b]{0.73\textwidth}
         \centering
         \includegraphics[width=\textwidth]{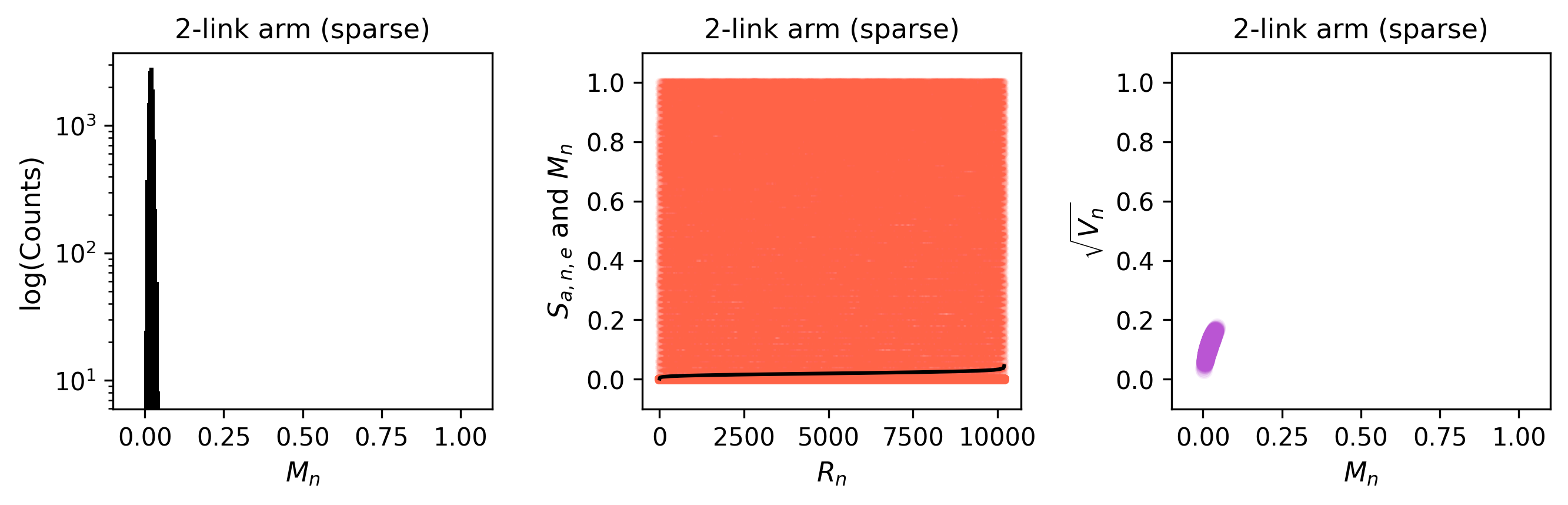}
         \caption{}
     \end{subfigure}
\caption{Performance distribution plots for the tasks: (a) \textit{1-link} \textit{(L=1.0m)}, (b) \textit{1-link} \textit{(L=1.65m)}, (c) \textit{2-link} arms with dense rewards, and (d) \textit{2-link} arm with sparse rewards. The left column shows a histogram of mean performances of the random policies (\textit{Log-scale histogram of} $M_{n}$). The middle column depicts \textit{mean performance curves} in black, i.e. mean performance $M_{n}$ vs rank $R_{n}$. Moreover, all the cumulative rewards of the policies $s_{a,n,e}$ across the trials are represented by red dots (behind the black curve). The right column displays plots of standard deviation $\sqrt{V_{n}}$ vs mean performance $M_{n}$ (often referred to as \textit{variance distribution}). The plots were made using $10^{4}$ random policies. }
\label{performance_plots}
\end{figure*}

Figure~\ref{performance_plots} presents the performance plots, where the left, middle and right columns, respectively, depict the \textit{mean performance histograms}, \textit{mean performance curves}, and \textit{variance distributions}. We normalised the mean $M_{n}$ and variance $V_{n}$ in each task to avoid scale-induced bias and ensure commensurability of performance~\citep{Cobbe20,Agarwal21}. We used \textit{min-max scaling}~\citep{Murphy12},
\begin{equation}\label{min-max}
    x' = \frac{x - \min\left[ x \right] }{ \max\left[ x \right] -  \min\left[ x \right] } 
\end{equation}   
where $x$ is the variable being scaled. 

\textbf{Overall description. }
We notice in Figure~\ref{performance_plots} that the histograms (across the tasks) have an overall shape that approximates a Gaussian distribution {that does not span the entire range of mean scores.} 
The performance curves (black curves) have smooth slopes without jumps, and {the variance distributions mostly reveal that performance consistency is not uniform across the mean score range.}
We discuss the plots in detail below.

% L = 1m vs 1.7m
\textbf{L = 1.0 vs L = 1.65m (\textit{1-link}, dense-rewards).}
In dense-reward settings for \textit{1-link} arms, where one has $L=1$ m and the other has $L=1.65$ m, we make the following observations,

\textit{Mean Performance histograms: } 
In both tasks, no single random policy achieved mean performance $M_{n}$ near the maximum return. This exhibits that the tasks are not trivial~\citep{Oller20}.
The performance distributions in both arms are similar. This shows that the tasks are structurally equivalent. This aligns with intuition, as the link lengths impacts reward scales (see Figure~\ref{RL_learning_across}), but do not alter the task structure, as seen after normalisation of rewards.  

\textit{Performance curves: }
Interestingly in both tasks, the random policies managed to attain episodic cumulative rewards $S_{n,e}$ that are nearly the maximum return (shown by red dots at $1.0$ ) in a few episodes. This is sensible since both the initial end-effector and target positions are random in every episode. With $500$ episodes, it is likely that initial and target positions of the end-effector were in close proximity in some episodes --- which simplifies the task in those episodes.

\textit{Variance distributions: }
{Both arms have similar variance distributions, with wide spread about the middle mean score (i.e. $~M_{n} = 0.5$), which slowly narrows towards the limits of the range of mean performance. This indicates that the majority of policies attain their mean performance by succeeding on some episodes and failing at others, leading to higher variance of scores across episodes. However, policies seem to consistently fail for lower mean scores and consistently succeed for higher mean scores, leading to lower variance~\citep{Oller20}.}
Note that apart from Figure~\ref{performance_plots} revealing that the \textit{1-link} arm tasks are similar, it is not clear which task is easier or harder between the two.

\textbf{1-Link vs 2-Link (dense rewards).}
Figure~\ref{performance_plots} ($2^{nd}$ and $3^{rd}$ rows) displays how the histogram has a narrower width for \textit{2-link} arm (than for \textit{1-link} arm) while the {median $M_{n}$ nearly remains} consistent across the tasks. The performance curve in \textit{2-link} arm has a low slope than that in the \textit{1-link} arm task. Furthermore, the variance of the performance in the \textit{2-link} arm task is smaller. These denote that the \textit{2-link} arm task is harder than the \textit{1-link} arm task. The reason is that harder tasks often provide higher rewards only when a coherent sequence of successful actions is executed, which untrained random policies are unlikely to achieve, hence reduced variability in performance.

\textbf{Dense vs Sparse rewards (2-link).}
Figure~\ref{performance_plots} ($3^{rd}$ and $4^{th}$ rows) portrays a drastic drop in peak $M_{n}$ and variance $V_n$, from dense- to sparse-reward settings. The performance curve slope further decreased (almost zero) in the sparse-reward setting compared to the dense-reward setting. This highlights a lack of diversity in the performance of the random policies. In the variance plots for the sparse-reward setting, we notice that random policies fail to succeed in the task regardless of initial conditions. We can conclude from these results that the dense-reward setting is easier than the sparse-reward setting.

\vspace{0.5em}
REMARK 2.~\textit{Although the statistical analysis and visualisation of performance provide some insights about the task characteristics and relative hardness, they fail to quantitatively measure task difficulty, i.e. the approach is qualitative. This makes it inapplicable to RL benchmarks and curriculum learning, where relative hardness amongst tasks needs to be quantified. Moreover, performance distributions that are similar across tasks can potentially make the plots less informative in comparing the tasks. For these reasons, we now examine quantitative metrics PIC and POIC.}
\vspace{0.5em}

\underline{\textsc{(II) PIC/POIC.}} 
The quantitative representations of task complexity offered by PIC and POIC are exhibited for our six tasks in Table~\ref{pic_vals}. We checked the statistical robustness of the results in Table~\ref{pic_vals} by quantifying their uncertainty using bootstrapped confidence intervals~\citep{Diciccio96}. These estimate the uncertainty by repeatedly resampling the data. In our context, the data are episodic cumulative rewards of random policies constructed via RWG.  We resampled the data $1,000$ times with replacement and computed the values presented in Table~\ref{pic_vals}. 

We then applied the Welch's t-test~\citep{Delacre17} to evaluate the statistical significance in the differences between values in Table~\ref{pic_vals}, using the same $1,000$ resamples. Consistently in all cases, the p-values of the t-statistic of the Welch's t-test are in the orders of $10^{-5}$, below a typical cut-off p-value $= 0.005$ which indicate strong statistical significance~\citep{Benjamin18}. Outcomes of the Welch's t-test and the confidence intervals of PIC and POIC can be found in Supplementary material\textsuperscript{\ref{fn:code1}}.
% Appendix~\ref{appx:stats_sig}.  

It should be noted that the values in Table~\ref{pic_vals} were gathered using a policy network of 2 hidden layers, each with 32 neurons. $10^{4}$ untrained policies were sampled from a multivariate normal prior distribution. We also confirmed that RWG sampled from different prior distributions and policy network architectures 
% (see Appendix~\ref{initialisation_methods})
did not change our results (see Supplementary materials\textsuperscript{\ref{fn:code1}}).   

\vspace{-0.5em}
\begin{table}[h!]
\caption{
PIC and POIC values with $N = 10^{4}$ samples (random policies). High PIC and POIC values correspond to easier tasks, while low values correspond to harder tasks.}
\vspace{-1.0em}
\centering
\begin{tabular}{| c | c | c | c |}
\hline
Rewards & Arm [dim] & PIC $\left(\times 10^{-3}\right)$ & POIC $\left(\times 10^{-3}\right)$ \\
\hline \hline
\multirow{2}{*}{Dense} & 1-link [1.0] & $4005\pm8.5$ & $2.628\pm0.056$ \\ %dense rewards
    & 1-link [1.65] & $4153\pm8.4$ & $4.105\pm0.085$ \\
    & 2-link [0.95,1.7] & $4200\pm6.1$ & $0.725\pm0.011$ \\
\hline
\multirow{3}{*}{Sparse} & 1-link [1.0] & $85.11\pm0.4$ & $1.958\pm0.034$ \\ %sparse rewards
    & 1-link [1.65] & $71.21\pm0.2$ & $1.197\pm0.031$ \\
    & 2-link [0.95,1.7] & $45.95\pm0.0$ & $0.946\pm0.0079$ \\    
\hline
\end{tabular}
\vspace{-0.5em}
\label{pic_vals}
\end{table}
% \vspace{-1.0em}

\textbf{Dense-reward settings.}
According to the PIC values under dense-reward settings (in Table~\ref{pic_vals}), the \textit{2-link} arm task is the easiest task (highest PIC) and \textit{1-link} arm ($L=1$ m) task is the hardest task (lowest PIC). This contradicts expectations based on Equations~\ref{task_difficulty_manipulator} and~\ref{error_bounds}, and our empirical RL results. For instance, we showed using learning curves of trained agents that the \textit{2-link} arm task is the hardest, while \textit{1-link} arm ($L=1$ m) task is the easiest. This is further corroborated by performance distributions of random policies in Figure~\ref{performance_plots}.  
On the POIC side, \textit{1-link} arm ($L=1.65$ m) task is easier than \textit{1-link} arm ($L=1$ m) task. This does not align with Equation~\ref{error_bounds}. 

\begin{figure*}[h!]
\centering
    \includegraphics[width=0.98\textwidth]{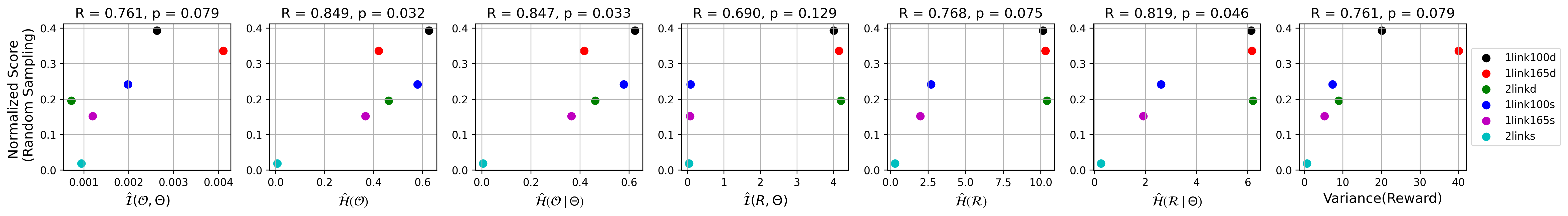} 
\vspace{-1em}
\caption{2D-scatter plots with Normalised scores (performance) computed using \textit{min-max scaling} (Equation~\ref{min-max}) over the returns of untrained random policies. The Normalised scores are plotted against PIC, POIC, variance of returns, along with entropies of optimality variable and cumulative reward (return) variable. 
}\label{2linkSvsD_plot} 
\end{figure*}

\textbf{Sparse-reward settings.}
In these settings, both the PIC and POIC order of task difficulty across the tasks seems correct. The tasks are ordered from easiest to hardest as \textit{1-link} arm ($L=1$ m), \textit{1-link} arm ($L=1.65$ m) and \textit{2-link} arm. This aligns with our intuition. 

When we compare across dense- and sparse-reward settings, POIC values for the \textit{2-link} arm task suggest that the dense-reward setting is harder than the sparse-reward setting. In this instance, POIC values contradict our expectations, as we showed empirically in Section~\ref{RL_Analysis} that dense-reward settings are easier than the sparse-reward settings. {This inconsistency of PIC and POIC is further observed in an additional task setting that places an obstacle in the workspace of the \textit{2-link} arm with dense rewards. Details are provided in Supplementary material\textsuperscript{\ref{fn:code1}} (Section D).}

To investigate these incorrect PIC and POIC instances, we decomposed the individual entropy terms in the metrics. Figure~\ref{2linkSvsD_plot} displays normalised scores (i.e. performance) against POIC, PIC, individual entropy terms, and the variance of cumulative rewards. The normalised scores use \textit{min-max scaling} (Equation~\ref{min-max}) over the performance samples of the random policies.  

\textbf{POIC related plots. }
The first three columns in Figure~\ref{2linkSvsD_plot} portray POIC and entropies of the optimality variable. We observe that $\hat{\mathcal{H}}(\mathcal{O})$ and $\hat{\mathcal{H}}(\mathcal{O} \mid \Theta)$ are closely approximate, which produces POIC $\hat{\mathcal{I}}(\mathcal{O};\Theta)$ values of small magnitude, similar to the work that introduced POIC~\citep{Furuta21}. There are multiple strong linear correlations between normalised scores and the other quantities (given by Pearson correlation coefficients above the plots), however they are not statistically significant.    

\textbf{PIC related and Variance plots. }
In Figure~\ref{2linkSvsD_plot}, the last four columns display PIC, entropies of the cumulative reward variable, and variance of returns. We note that $\hat{\mathcal{H}}({R})$ and $\hat{\mathcal{H}}({R} \mid \Theta)$ differ. This is responsible for larger magnitude values of PIC $\hat{\mathcal{I}}({R};\Theta)$. It seems dense-reward settings have more variability in returns, than sparse-reward settings. This aligns with results presented in  Figure~\ref{performance_plots}. It seems generally that in our setup, tasks with dense-rewards enjoy higher normalised scores than sparse-rewards coinciding with our expectations. Figure~\ref{2linkSvsD_plot} does not provide further insights about why PIC and POIC values in Table~\ref{pic_vals} do not match expectations. We discuss further possibilities in the next section.
\vspace{-1.0em}

%%%%%%%%%%%%%%%%%%%%%%%%%%%%%%%%%%%%%%%%%%%%%%%%%%%%%%%%%%%%%%%%%%%%%%%%
\section{Discussion \& Limitations}\label{tx:discussion}
Finally, we explore potential reasons for the inconsistencies observed with RWG-based {metrics} PIC and POIC compared to expectations derived from robotic control and verified with empirical RL. These issues stem from (1) the dependence on randomly generated parameters on the prior distribution $p(\theta)$, and (2) the lack of consideration for training and exploration. 
%We outline directions for future investigations.

While statistical analysis of performance of RWG-generated policies has been consistent with {intuition}, it can be challenging to effectively communicate the degree of difference in task difficulty. Moreover, if tasks produce nearly similar performance distributions, then this approach might be less informative for comparative analysis. 

%RWG limitation resurfacing
PIC and POIC are dependent on the prior distribution of parameters $p(\theta)$, as highlighted in~\citep{Furuta21}. The prior $p(\theta)$ can be interpreted as the \textit{effective search area} in the parameter (policy) space~\citep{Aleksandrowicz23}. For problems where high-performing policies are sparsely distributed in the parameter space, the effective search area is likely to cover mainly {low-performing} 
regions. {Policies sampled from these regions are limited to yielding mean performance $M_n$ far from the peak return as shown in Figure~\ref{performance_plots}}.
This observation reinforces the limitation of RWG originally noted by~\cite{Schmidhuber99}, namely its ineffectiveness in tasks with sparse solution regions in the weight space.

%Lack of exploration
Another limitation arises from the fact that RWG does not involve training. This implies that the effective search area remains static once $p(\theta)$ is selected. In contrast, training involves exploration of the policy space~\citep{Sutton18,Ladosz22,Nkhumise25} --- where the effective search area (of the learning algorithm) is dynamically moved around in the policy space. It is also important to note that neither statistical analysis of the performance of random policies, nor PIC and POIC metrics, consider the visitation complexity~\citep{Conserva22} of the tasks, which measures the difficulty in exploring the state space of the environment. This implies that the way actions influence state transitions during exploration in the learning phase is not accounted for by these task complexity methods~\citep{Furuta21}. Several methods that aim to capture exploration effort~\citep{Amin21,Ladosz22,Nkhumise25} have been investigated; however, none have been applied to task complexity. This makes for an interesting future direction of work.

It is clear from our experimental results that PIC and POIC can be misleading in capturing task complexity. Inconsistencies in these metrics can be challenging to notice in most RL benchmarks, especially if they have tasks with heterogeneous structure. Our task framework offers tasks with structural homogeneity and known relative task complexity, thus enabling a more reliable assessment of these metrics.  

The results presented in this article showcase the need for continued work in task complexity for deep RL, especially for the case of robotic tasks, where our task framework could be a starting point. We propose the following directions for improving PIC and POIC:

% {\begin{enumerate}
% \textcolor{red}
{(1)  A key drawback of the PIC/POIC metrics is the limited effective search space sampled by RWG. We can replace the standard multilayer perceptron (MLP) policy with an architecture which introduces parameterised inductive biases relevant to the tasks. These biases are aimed at maximally covering the state-action space. By setting the parameters of these inductive biases via RWG, we can cover a wider search space. In robotics, the inductive bias can be a policy composed of dynamic movement primitives~\citep{Stulp11}, skills~\citep{Dalal21} or normalising flows~\citep{Khader21}. The drawback of this approach is that the resulting complexity measure would be dependent on the selected inductive biases.}

% \textcolor{red}
{(2) A second issue with PIC and POIC metrics is that they have a static effective search area in the parameter space due to the lack of exploration. In devising new metrics, we want the effective search area to be dynamic by including exploration. We can perform RWG after every $k$ updates of a RL algorithm and compute corresponding PIC/POIC values at each stage. Ultimately, use the mean PIC/POIC values across the entire learning path as the metric for task complexity. This would entail sampling weights via RWG, at every k update of the policy during RL training $\theta_n$ using $\theta_n = \theta_k + \epsilon$, $\epsilon \sim \mathcal{N}(0,I)$ where $\theta_k$ are parameters after the $k^{th}$ policy model update. With every RWG set, we compute $PIC_k$, and finally,  determine the average across the trajectory and use it as a measure of task complexity. The drawback of this approach is that the task complexity metric(s) would be dependent on the exploration strategy of the learning algorithm.}  

% \textcolor{red}{
(3) When the optimal policy is known, we can compute the distribution of distances of any RWG policy to the optimal policy using optimal transport~\citep{Nkhumise25}. Task complexity could then be defined by the mean and variance of these distances, thereby capturing the expected effort required to move from random policies to the optimal policy. The limitation of this method is that it requires prior knowledge of the optimal policy. 

In principle, all three of these suggestions could be combined.
% \end{enumerate}

%%%%%%%%%%%%%%%%%%%%%%%%%%%%%%%%%%%%%%%%%%%%%%%%%%%%%%%%%%%%%%%%%%%%%%%%

%%% The acknowledgments section is defined using the "acks" environment
%%% (rather than an unnumbered section). The use of this environment 
%%% ensures the proper identification of the section in the article 
%%% metadata as well as the consistent spelling of the heading.

\begin{acks}
R. Nkhumise was supported by the EPSRC Doctoral Training Partnership (DTP) - Early Career Researcher funding awarded to A. Gilra. A. Gilra acknowledges the CHIST-ERA grant for the Causal Explanations in Reinforcement Learning (CausalXRL) project (CHIST-ERA-19-XAI-002), by the Engineering and Physical Sciences Research Council, United Kingdom (grant reference EP/V055720/1) for supporting the work.
\end{acks}

%%%%%%%%%%%%%%%%%%%%%%%%%%%%%%%%%%%%%%%%%%%%%%%%%%%%%%%%%%%%%%%%%%%%%%%%

%%% The next two lines define, first, the bibliography style to be 
%%% applied, and, second, the bibliography file to be used.

% \bibliographystyle{ACM-Reference-Format} 
% \bibliography{references}

\begin{thebibliography}{64}

%%% the Web address.

\ifx \showCODEN    \undefined \def \showCODEN     #1{\unskip}     \fi
\ifx \showDOI      \undefined \def \showDOI       #1{#1}\fi
\ifx \showISBNx    \undefined \def \showISBNx     #1{\unskip}     \fi
\ifx \showISBNxiii \undefined \def \showISBNxiii  #1{\unskip}     \fi
\ifx \showISSN     \undefined \def \showISSN      #1{\unskip}     \fi
\ifx \showLCCN     \undefined \def \showLCCN      #1{\unskip}     \fi
\ifx \shownote     \undefined \def \shownote      #1{#1}          \fi
\ifx \showarticletitle \undefined \def \showarticletitle #1{#1}   \fi
\ifx \showURL      \undefined \def \showURL       {\relax}        \fi
% invisible to TeX
\providecommand\bibfield[2]{#2}
\providecommand\bibinfo[2]{#2}
\providecommand\natexlab[1]{#1}
\providecommand\showeprint[2][]{arXiv:#2}

\bibitem[\protect\citeauthoryear{Abel, Allen, Arumugam, Hershkowitz, Littman, and Wong}{Abel et~al\mbox{.}}{2021}]%
        {Abel21}
\bibfield{author}{\bibinfo{person}{D. Abel}, \bibinfo{person}{C. Allen}, \bibinfo{person}{D. Arumugam}, \bibinfo{person}{D.~E. Hershkowitz}, \bibinfo{person}{M.~L. Littman}, {and} \bibinfo{person}{L.~L.~S. Wong}.} \bibinfo{year}{2021}\natexlab{}.
\newblock \showarticletitle{Bad-policy density: A measure of reinforcement learning hardness}.
\newblock \bibinfo{journal}{\emph{arXiv preprint arXiv:2110.03424}} (\bibinfo{year}{2021}).
\newblock
\showeprint[arxiv]{2110.03424}


\bibitem[\protect\citeauthoryear{Agarwal, Schwarzer, Castro, Courville, and Bellemare}{Agarwal et~al\mbox{.}}{2021}]%
        {Agarwal21}
\bibfield{author}{\bibinfo{person}{R. Agarwal}, \bibinfo{person}{M. Schwarzer}, \bibinfo{person}{P.~S. Castro}, \bibinfo{person}{A.~C. Courville}, {and} \bibinfo{person}{M. Bellemare}.} \bibinfo{year}{2021}\natexlab{}.
\newblock \showarticletitle{Deep reinforcement learning at the edge of the statistical precipice}.
\newblock \bibinfo{journal}{\emph{Advances in neural information processing systems}}  \bibinfo{volume}{34} (\bibinfo{year}{2021}), \bibinfo{pages}{29304--29320}.
\newblock


\bibitem[\protect\citeauthoryear{Aleksandrowicz and Jaworek-Korjakowska}{Aleksandrowicz and Jaworek-Korjakowska}{2023}]%
        {Aleksandrowicz23}
\bibfield{author}{\bibinfo{person}{M. Aleksandrowicz} {and} \bibinfo{person}{J. Jaworek-Korjakowska}.} \bibinfo{year}{2023}\natexlab{}.
\newblock \showarticletitle{Metrics for assessing generalization of deep reinforcement learning in parameterized environments}.
\newblock \bibinfo{journal}{\emph{JAISCR}} \bibinfo{volume}{14}, \bibinfo{number}{1} (\bibinfo{year}{2023}), \bibinfo{pages}{45--61}.
\newblock


\bibitem[\protect\citeauthoryear{Amin, Gomrokchi, Satija, van Hoof, and Precup}{Amin et~al\mbox{.}}{2021}]%
        {Amin21}
\bibfield{author}{\bibinfo{person}{S. Amin}, \bibinfo{person}{M. Gomrokchi}, \bibinfo{person}{H. Satija}, \bibinfo{person}{H. van Hoof}, {and} \bibinfo{person}{D. Precup}.} \bibinfo{year}{2021}\natexlab{}.
\newblock \showarticletitle{A Survey of Exploration Methods in Reinforcement Learning}.
\newblock \bibinfo{journal}{\emph{arXiv preprint arXiv:2109.00157}} (\bibinfo{year}{2021}).
\newblock
\showeprint[arxiv]{2109.00157}


\bibitem[\protect\citeauthoryear{Andrychowicz, Wolski, Ray, Schneider, Fong, Welinder, McGrew, Tobin, A.~Pieter~Abbeel, and Zaremba}{Andrychowicz et~al\mbox{.}}{2017}]%
        {Andrychowicz17}
\bibfield{author}{\bibinfo{person}{M. Andrychowicz}, \bibinfo{person}{F. Wolski}, \bibinfo{person}{A. Ray}, \bibinfo{person}{J. Schneider}, \bibinfo{person}{R. Fong}, \bibinfo{person}{P. Welinder}, \bibinfo{person}{B. McGrew}, \bibinfo{person}{J. Tobin}, \bibinfo{person}{OpenAI A.~Pieter~Abbeel}, {and} \bibinfo{person}{W. Zaremba}.} \bibinfo{year}{2017}\natexlab{}.
\newblock \showarticletitle{Hindsight Experience Replay}. In \bibinfo{booktitle}{\emph{Advances in Neural Information Processing Systems}}, Vol.~\bibinfo{volume}{30}. \bibinfo{publisher}{Curran Associates, Inc.}
\newblock


\bibitem[\protect\citeauthoryear{Bellemare, Naddaf, Veness, and Bowling}{Bellemare et~al\mbox{.}}{2013}]%
        {Bellemare13}
\bibfield{author}{\bibinfo{person}{M.~G. Bellemare}, \bibinfo{person}{Y. Naddaf}, \bibinfo{person}{J. Veness}, {and} \bibinfo{person}{M. Bowling}.} \bibinfo{year}{2013}\natexlab{}.
\newblock \showarticletitle{The arcade learning environment: An evaluation platform for general agents}.
\newblock \bibinfo{journal}{\emph{Journal of artificial intelligence research}}  \bibinfo{volume}{47} (\bibinfo{year}{2013}), \bibinfo{pages}{253--279}.
\newblock


\bibitem[\protect\citeauthoryear{Bengio, Louradour, Collobert, and Weston}{Bengio et~al\mbox{.}}{2009}]%
        {Bengio09}
\bibfield{author}{\bibinfo{person}{Y. Bengio}, \bibinfo{person}{J. Louradour}, \bibinfo{person}{R. Collobert}, {and} \bibinfo{person}{J. Weston}.} \bibinfo{year}{2009}\natexlab{}.
\newblock \showarticletitle{Curriculum learning}. In \bibinfo{booktitle}{\emph{Proceedings of the 26th annual international conference on machine learning}}. \bibinfo{pages}{41--48}.
\newblock


\bibitem[\protect\citeauthoryear{Benjamin, Berger, Johannesson, Nosek, Wagenmakers, Berk, Bollen, Brembs, Brown, Camerer, et~al\mbox{.}}{Benjamin et~al\mbox{.}}{2018}]%
        {Benjamin18}
\bibfield{author}{\bibinfo{person}{D.~J. Benjamin}, \bibinfo{person}{J.~O. Berger}, \bibinfo{person}{M. Johannesson}, \bibinfo{person}{B.~A. Nosek}, \bibinfo{person}{E.~J. Wagenmakers}, \bibinfo{person}{R. Berk}, \bibinfo{person}{K.~A. Bollen}, \bibinfo{person}{B. Brembs}, \bibinfo{person}{L. Brown}, \bibinfo{person}{C. Camerer}, {et~al\mbox{.}}} \bibinfo{year}{2018}\natexlab{}.
\newblock \showarticletitle{Redefine statistical significance}.
\newblock \bibinfo{journal}{\emph{Nature human behaviour}} \bibinfo{volume}{2}, \bibinfo{number}{1} (\bibinfo{year}{2018}), \bibinfo{pages}{6--10}.
\newblock


\bibitem[\protect\citeauthoryear{Brockman, Cheung, Pettersson, Schneider, Schulman, Tang, and Zaremba}{Brockman et~al\mbox{.}}{2016}]%
        {Brockman16}
\bibfield{author}{\bibinfo{person}{G. Brockman}, \bibinfo{person}{V. Cheung}, \bibinfo{person}{L. Pettersson}, \bibinfo{person}{J. Schneider}, \bibinfo{person}{J. Schulman}, \bibinfo{person}{J. Tang}, {and} \bibinfo{person}{W. Zaremba}.} \bibinfo{year}{2016}\natexlab{}.
\newblock \showarticletitle{OpenAI Gym}.
\newblock \bibinfo{journal}{\emph{arXiv preprint arXiv:1606.01540}} (\bibinfo{year}{2016}).
\newblock
\showeprint[arxiv]{1606.01540}
\urldef\tempurl%
\url{https://www.gymlibrary.dev/environments/classic_control/mountain_car/}
\showURL{%
\tempurl}


\bibitem[\protect\citeauthoryear{Chiaverini}{Chiaverini}{2002}]%
        {Chiaverini02}
\bibfield{author}{\bibinfo{person}{S. Chiaverini}.} \bibinfo{year}{2002}\natexlab{}.
\newblock \showarticletitle{Singularity-robust task-priority redundancy resolution for real-time kinematic control of robot manipulators}.
\newblock \bibinfo{journal}{\emph{IEEE Transactions on Robotics and Automation}} \bibinfo{volume}{13}, \bibinfo{number}{3} (\bibinfo{year}{2002}), \bibinfo{pages}{398--410}.
\newblock


\bibitem[\protect\citeauthoryear{Cobbe, Hesse, Hilton, and Schulman}{Cobbe et~al\mbox{.}}{2020}]%
        {Cobbe20}
\bibfield{author}{\bibinfo{person}{K. Cobbe}, \bibinfo{person}{C. Hesse}, \bibinfo{person}{J. Hilton}, {and} \bibinfo{person}{J. Schulman}.} \bibinfo{year}{2020}\natexlab{}.
\newblock \showarticletitle{Leveraging procedural generation to benchmark reinforcement learning}. In \bibinfo{booktitle}{\emph{International conference on machine learning}}.
\newblock


\bibitem[\protect\citeauthoryear{Cobbe, Klimov, Hesse, Kim, and Schulman}{Cobbe et~al\mbox{.}}{2019}]%
        {Cobbe19}
\bibfield{author}{\bibinfo{person}{K. Cobbe}, \bibinfo{person}{O. Klimov}, \bibinfo{person}{C. Hesse}, \bibinfo{person}{T. Kim}, {and} \bibinfo{person}{J. Schulman}.} \bibinfo{year}{2019}\natexlab{}.
\newblock \showarticletitle{Quantifying generalization in reinforcement learning}. In \bibinfo{booktitle}{\emph{International conference on machine learning}}.
\newblock


\bibitem[\protect\citeauthoryear{Conserva and Rauber}{Conserva and Rauber}{2022}]%
        {Conserva22}
\bibfield{author}{\bibinfo{person}{M. Conserva} {and} \bibinfo{person}{P. Rauber}.} \bibinfo{year}{2022}\natexlab{}.
\newblock \showarticletitle{Hardness in Markov Decision Processes: Theory and Practice}.
\newblock \bibinfo{journal}{\emph{36th Conference on Neural Information Processing Systems}} (\bibinfo{year}{2022}).
\newblock


\bibitem[\protect\citeauthoryear{Conserva, Sasso, and Rauber}{Conserva et~al\mbox{.}}{2025}]%
        {Conserva25}
\bibfield{author}{\bibinfo{person}{M. Conserva}, \bibinfo{person}{R. Sasso}, {and} \bibinfo{person}{P. Rauber}.} \bibinfo{year}{2025}\natexlab{}.
\newblock \showarticletitle{On the Limits of Tabular Hardness Metrics for Deep RL: A Study with the Pharos Benchmark}.
\newblock \bibinfo{journal}{\emph{arXiv preprint arXiv:2509.17092}} (\bibinfo{year}{2025}).
\newblock
\showeprint[arxiv]{2509.17092}


\bibitem[\protect\citeauthoryear{Copot, Muresan, Ionescu, Vanlanduit, and Keyser}{Copot et~al\mbox{.}}{2018}]%
        {Copot18}
\bibfield{author}{\bibinfo{person}{C. Copot}, \bibinfo{person}{C. Muresan}, \bibinfo{person}{C.-M. Ionescu}, \bibinfo{person}{S. Vanlanduit}, {and} \bibinfo{person}{R.~De Keyser}.} \bibinfo{year}{2018}\natexlab{}.
\newblock \showarticletitle{Calibration of UR10 robot controller through simple auto-tuning approach}.
\newblock \bibinfo{journal}{\emph{Robotics}} \bibinfo{volume}{7}, \bibinfo{number}{3} (\bibinfo{year}{2018}).
\newblock
\urldef\tempurl%
\url{https://www.mdpi.com/2218-6581/7/3/35}
\showURL{%
\tempurl}


\bibitem[\protect\citeauthoryear{Corke}{Corke}{2011}]%
        {Corke11}
\bibfield{author}{\bibinfo{person}{P. Corke}.} \bibinfo{year}{2011}\natexlab{}.
\newblock \bibinfo{booktitle}{\emph{Robotics, vision and control}}.
\newblock \bibinfo{publisher}{Springer Berlin, Heidelberg}.
\newblock


\bibitem[\protect\citeauthoryear{Cover and Thomas}{Cover and Thomas}{2006}]%
        {Cover06}
\bibfield{author}{\bibinfo{person}{T.~M. Cover} {and} \bibinfo{person}{J.~A. Thomas}.} \bibinfo{year}{2006}\natexlab{}.
\newblock \bibinfo{booktitle}{\emph{Elements of Information Theory} (\bibinfo{edition}{2nd} ed.)}.
\newblock \bibinfo{publisher}{John Wiley \& Sons, Ltd}.
\newblock


\bibitem[\protect\citeauthoryear{Dalal, Pathak, and Salakhutdinov}{Dalal et~al\mbox{.}}{2021}]%
        {Dalal21}
\bibfield{author}{\bibinfo{person}{M. Dalal}, \bibinfo{person}{D. Pathak}, {and} \bibinfo{person}{R.~R. Salakhutdinov}.} \bibinfo{year}{2021}\natexlab{}.
\newblock \showarticletitle{Accelerating robotic reinforcement learning via parameterized action primitives}.
\newblock \bibinfo{journal}{\emph{Advances in Neural Information Processing Systems}}  \bibinfo{volume}{34} (\bibinfo{year}{2021}), \bibinfo{pages}{21847--21859}.
\newblock


\bibitem[\protect\citeauthoryear{Delacre, Lakens, and Leys}{Delacre et~al\mbox{.}}{2017}]%
        {Delacre17}
\bibfield{author}{\bibinfo{person}{M. Delacre}, \bibinfo{person}{D. Lakens}, {and} \bibinfo{person}{C. Leys}.} \bibinfo{year}{2017}\natexlab{}.
\newblock \showarticletitle{Why psychologists should by default use Welch's t-test instead of Student's t-test}.
\newblock \bibinfo{journal}{\emph{International Review of Social Psychology}} \bibinfo{volume}{30}, \bibinfo{number}{1} (\bibinfo{year}{2017}), \bibinfo{pages}{92--101}.
\newblock


\bibitem[\protect\citeauthoryear{DiCiccio and Efron}{DiCiccio and Efron}{1996}]%
        {Diciccio96}
\bibfield{author}{\bibinfo{person}{T.~J. DiCiccio} {and} \bibinfo{person}{B. Efron}.} \bibinfo{year}{1996}\natexlab{}.
\newblock \showarticletitle{Bootstrap confidence intervals}.
\newblock \bibinfo{journal}{\emph{Statistical science}} \bibinfo{volume}{11}, \bibinfo{number}{3} (\bibinfo{year}{1996}), \bibinfo{pages}{189--228}.
\newblock


\bibitem[\protect\citeauthoryear{Duan, Chen, Houthooft, Schulman, and Abbeel}{Duan et~al\mbox{.}}{2016}]%
        {Duan16}
\bibfield{author}{\bibinfo{person}{Y. Duan}, \bibinfo{person}{X. Chen}, \bibinfo{person}{R. Houthooft}, \bibinfo{person}{J. Schulman}, {and} \bibinfo{person}{P. Abbeel}.} \bibinfo{year}{2016}\natexlab{}.
\newblock \showarticletitle{Benchmarking Deep Reinforcement Learning for Continuous Control}.
\newblock \bibinfo{journal}{\emph{Proceedings of the 33 rd International Conference on Machine Learning}}  \bibinfo{volume}{48} (\bibinfo{year}{2016}), \bibinfo{pages}{1329--–1338}.
\newblock


\bibitem[\protect\citeauthoryear{Dulac-Arnold, Levine, Mankowitz, Li, Paduraru, Gowal, and Hester}{Dulac-Arnold et~al\mbox{.}}{2020}]%
        {Dulac20}
\bibfield{author}{\bibinfo{person}{G. Dulac-Arnold}, \bibinfo{person}{N. Levine}, \bibinfo{person}{D.~J. Mankowitz}, \bibinfo{person}{J. Li}, \bibinfo{person}{C. Paduraru}, \bibinfo{person}{S. Gowal}, {and} \bibinfo{person}{T. Hester}.} \bibinfo{year}{2020}\natexlab{}.
\newblock \showarticletitle{An empirical investigation of the challenges of real-world reinforcement learning}.
\newblock \bibinfo{journal}{\emph{arXiv preprint arXiv:2003.11881}} (\bibinfo{year}{2020}).
\newblock
\showeprint[arxiv]{2003.11881}


\bibitem[\protect\citeauthoryear{Edmondson and Petrick}{Edmondson and Petrick}{2025}]%
        {Edmondson25}
\bibfield{author}{\bibinfo{person}{A. Edmondson} {and} \bibinfo{person}{R.~P.~A. Petrick}.} \bibinfo{year}{2025}\natexlab{}.
\newblock \showarticletitle{Navigating Errors: The Tolerance of Reinforcement Learning Algorithms to Misleading Heuristics}.
\newblock \bibinfo{journal}{\emph{Association for the Advancement of Artificial Intelligence}} (\bibinfo{year}{2025}).
\newblock


\bibitem[\protect\citeauthoryear{Featherstone}{Featherstone}{2008}]%
        {Featherstone08}
\bibfield{author}{\bibinfo{person}{R. Featherstone}.} \bibinfo{year}{2008}\natexlab{}.
\newblock \bibinfo{booktitle}{\emph{Rigid body dynamics algorithms}}.
\newblock \bibinfo{publisher}{Springer}.
\newblock


\bibitem[\protect\citeauthoryear{Furuta, Matsushima, Kozuno, Matsuo, Levine, Nachum, and Gu}{Furuta et~al\mbox{.}}{2021}]%
        {Furuta21}
\bibfield{author}{\bibinfo{person}{H. Furuta}, \bibinfo{person}{T. Matsushima}, \bibinfo{person}{T. Kozuno}, \bibinfo{person}{Y. Matsuo}, \bibinfo{person}{S. Levine}, \bibinfo{person}{O. Nachum}, {and} \bibinfo{person}{S.~S. Gu}.} \bibinfo{year}{2021}\natexlab{}.
\newblock \showarticletitle{Policy Information Capacity: Information-Theoretic Measure for Task Complexity in Deep Reinforcement Learning}.
\newblock \bibinfo{journal}{\emph{Proceedings of the 38th International Conference on Machine Learning}}  \bibinfo{volume}{139} (\bibinfo{year}{2021}), \bibinfo{pages}{3541--3552}.
\newblock


\bibitem[\protect\citeauthoryear{Haarnoja, Zhou, Abbeel, and Levine}{Haarnoja et~al\mbox{.}}{2018}]%
        {Haarnoja18}
\bibfield{author}{\bibinfo{person}{T. Haarnoja}, \bibinfo{person}{A. Zhou}, \bibinfo{person}{P. Abbeel}, {and} \bibinfo{person}{S. Levine}.} \bibinfo{year}{2018}\natexlab{}.
\newblock \showarticletitle{Soft Actor-Critic: Off-Policy Maximum Entropy Deep Reinforcement Learning with a Stochastic Actor}. In \bibinfo{booktitle}{\emph{Proceedings of the 35th International Conference on Machine Learning}}.
\newblock


\bibitem[\protect\citeauthoryear{Hentout, Maoudj, and Aouache}{Hentout et~al\mbox{.}}{2023}]%
        {Hentout23}
\bibfield{author}{\bibinfo{person}{A. Hentout}, \bibinfo{person}{A. Maoudj}, {and} \bibinfo{person}{M. Aouache}.} \bibinfo{year}{2023}\natexlab{}.
\newblock \showarticletitle{A review of the literature on fuzzy-logic approaches for collision-free path planning of manipulator robots}.
\newblock \bibinfo{journal}{\emph{Artificial Intelligence Review}} \bibinfo{volume}{56}, \bibinfo{number}{4} (\bibinfo{year}{2023}), \bibinfo{pages}{3369--3444}.
\newblock


\bibitem[\protect\citeauthoryear{Ibarz, Tan, Finn, Kalakrishnan, Pastor, and Levine}{Ibarz et~al\mbox{.}}{2021}]%
        {Ibarz21}
\bibfield{author}{\bibinfo{person}{J. Ibarz}, \bibinfo{person}{J. Tan}, \bibinfo{person}{C. Finn}, \bibinfo{person}{M. Kalakrishnan}, \bibinfo{person}{P. Pastor}, {and} \bibinfo{person}{S. Levine}.} \bibinfo{year}{2021}\natexlab{}.
\newblock \showarticletitle{How to train your robot with deep reinforcement learning: lessons we have learned}.
\newblock \bibinfo{journal}{\emph{The International Journal of Robotics Research}} \bibinfo{volume}{40}, \bibinfo{number}{4-5} (\bibinfo{year}{2021}), \bibinfo{pages}{698--721}.
\newblock


\bibitem[\protect\citeauthoryear{Jaderberg, Czarnecki, Dunning, Marris, Lever, Castaneda, Beattie, Rabinowitz, Morcos, Ruderman, Sonnerat, Green, Deason, Leibo, Silver, Hassabis, Kavukcuoglu, and Graepel}{Jaderberg et~al\mbox{.}}{2019}]%
        {Jaderberg19}
\bibfield{author}{\bibinfo{person}{M. Jaderberg}, \bibinfo{person}{W.~M. Czarnecki}, \bibinfo{person}{I. Dunning}, \bibinfo{person}{L. Marris}, \bibinfo{person}{G. Lever}, \bibinfo{person}{A.~G. Castaneda}, \bibinfo{person}{C. Beattie}, \bibinfo{person}{N.~C. Rabinowitz}, \bibinfo{person}{A.~S. Morcos}, \bibinfo{person}{A. Ruderman}, \bibinfo{person}{N. Sonnerat}, \bibinfo{person}{T. Green}, \bibinfo{person}{L. Deason}, \bibinfo{person}{J.~Z. Leibo}, \bibinfo{person}{D. Silver}, \bibinfo{person}{D. Hassabis}, \bibinfo{person}{K. Kavukcuoglu}, {and} \bibinfo{person}{T. Graepel}.} \bibinfo{year}{2019}\natexlab{}.
\newblock \showarticletitle{Human-level performance in 3D multiplayer games with population-based reinforcement learning}.
\newblock \bibinfo{journal}{\emph{Science}} \bibinfo{volume}{364}, \bibinfo{number}{6443} (\bibinfo{year}{2019}), \bibinfo{pages}{859--865}.
\newblock


\bibitem[\protect\citeauthoryear{Justesen, Torrado, Bontrager, Khalifa, Togelius, and Risi}{Justesen et~al\mbox{.}}{2018}]%
        {Justesen18}
\bibfield{author}{\bibinfo{person}{N. Justesen}, \bibinfo{person}{R.~R. Torrado}, \bibinfo{person}{P. Bontrager}, \bibinfo{person}{A. Khalifa}, \bibinfo{person}{J. Togelius}, {and} \bibinfo{person}{S. Risi}.} \bibinfo{year}{2018}\natexlab{}.
\newblock \showarticletitle{Illuminating generalization in deep reinforcement learning through procedural level generation}.
\newblock \bibinfo{journal}{\emph{arXiv preprint arXiv:1806.10729}} (\bibinfo{year}{2018}).
\newblock
\showeprint[arxiv]{1806.10729}


\bibitem[\protect\citeauthoryear{Khadem, Cruz, and Bergeles}{Khadem et~al\mbox{.}}{2018}]%
        {Khadem18}
\bibfield{author}{\bibinfo{person}{M. Khadem}, \bibinfo{person}{L.~Da Cruz}, {and} \bibinfo{person}{C. Bergeles}.} \bibinfo{year}{2018}\natexlab{}.
\newblock \showarticletitle{Force/velocity manipulability analysis for 3d continuum robots}. In \bibinfo{booktitle}{\emph{2018 IEEE/RSJ International Conference on Intelligent Robots and Systems (IROS)}}. IEEE, \bibinfo{pages}{4920--4926}.
\newblock


\bibitem[\protect\citeauthoryear{Khader, Yin, Falco, and Kragic}{Khader et~al\mbox{.}}{2021}]%
        {Khader21}
\bibfield{author}{\bibinfo{person}{S.~A. Khader}, \bibinfo{person}{H. Yin}, \bibinfo{person}{P. Falco}, {and} \bibinfo{person}{D. Kragic}.} \bibinfo{year}{2021}\natexlab{}.
\newblock \showarticletitle{Learning stable normalizing-flow control for robotic manipulation}. In \bibinfo{booktitle}{\emph{2021 IEEE International Conference on Robotics and Automation (ICRA)}}. IEEE, \bibinfo{pages}{1644--1650}.
\newblock


\bibitem[\protect\citeauthoryear{Kingma and Ba}{Kingma and Ba}{2017}]%
        {Kingma17}
\bibfield{author}{\bibinfo{person}{D.~P. Kingma} {and} \bibinfo{person}{J. Ba}.} \bibinfo{year}{2017}\natexlab{}.
\newblock \showarticletitle{Adam: A Method for Stochastic Optimization}.
\newblock \bibinfo{journal}{\emph{arXiv preprint arXiv:1412.6980}} (\bibinfo{year}{2017}).
\newblock
\showeprint[arxiv]{1412.6980}


\bibitem[\protect\citeauthoryear{Kormushev, Calinon, and Caldwell}{Kormushev et~al\mbox{.}}{2013}]%
        {Kormushev13}
\bibfield{author}{\bibinfo{person}{P. Kormushev}, \bibinfo{person}{S. Calinon}, {and} \bibinfo{person}{D.~G. Caldwell}.} \bibinfo{year}{2013}\natexlab{}.
\newblock \showarticletitle{Reinforcement learning in robotics: Applications and real-world challenges}.
\newblock \bibinfo{journal}{\emph{Robotics}} \bibinfo{volume}{2}, \bibinfo{number}{3} (\bibinfo{year}{2013}), \bibinfo{pages}{122--148}.
\newblock


\bibitem[\protect\citeauthoryear{Ladosz, Weng, Kim, and Oh}{Ladosz et~al\mbox{.}}{2022}]%
        {Ladosz22}
\bibfield{author}{\bibinfo{person}{P. Ladosz}, \bibinfo{person}{L. Weng}, \bibinfo{person}{M. Kim}, {and} \bibinfo{person}{H. Oh}.} \bibinfo{year}{2022}\natexlab{}.
\newblock \showarticletitle{Exploration in Deep Reinforcement Learning: A Survey}.
\newblock \bibinfo{journal}{\emph{Information Fusion}}  \bibinfo{volume}{85} (\bibinfo{year}{2022}), \bibinfo{pages}{1--22}.
\newblock


\bibitem[\protect\citeauthoryear{Mason}{Mason}{2018}]%
        {Mason18}
\bibfield{author}{\bibinfo{person}{M.~T. Mason}.} \bibinfo{year}{2018}\natexlab{}.
\newblock \showarticletitle{Toward robotic manipulation}.
\newblock \bibinfo{journal}{\emph{Annual Review of Control, Robotics, and Autonomous Systems}} \bibinfo{volume}{1}, \bibinfo{number}{1} (\bibinfo{year}{2018}), \bibinfo{pages}{1--28}.
\newblock


\bibitem[\protect\citeauthoryear{Murphy}{Murphy}{2012}]%
        {Murphy12}
\bibfield{author}{\bibinfo{person}{K.~P. Murphy}.} \bibinfo{year}{2012}\natexlab{}.
\newblock \bibinfo{booktitle}{\emph{Machine learning: a probabilistic perspective}}.
\newblock \bibinfo{publisher}{The MIT Press}.
\newblock
\showISBNx{9780262018029}


\bibitem[\protect\citeauthoryear{Murray, Li, and Sastry}{Murray et~al\mbox{.}}{2017}]%
        {Murray17}
\bibfield{author}{\bibinfo{person}{R.~M. Murray}, \bibinfo{person}{Z. Li}, {and} \bibinfo{person}{S.~S. Sastry}.} \bibinfo{year}{2017}\natexlab{}.
\newblock \bibinfo{booktitle}{\emph{A mathematical introduction to robotic manipulation}}.
\newblock \bibinfo{publisher}{CRC press}.
\newblock


\bibitem[\protect\citeauthoryear{Nakanishi, Cory, Mistry, Peters, and Schaal}{Nakanishi et~al\mbox{.}}{2008}]%
        {Nakanishi08}
\bibfield{author}{\bibinfo{person}{J. Nakanishi}, \bibinfo{person}{R. Cory}, \bibinfo{person}{M. Mistry}, \bibinfo{person}{J. Peters}, {and} \bibinfo{person}{S. Schaal}.} \bibinfo{year}{2008}\natexlab{}.
\newblock \showarticletitle{Operational space control: A theoretical and empirical comparison}.
\newblock \bibinfo{journal}{\emph{The International Journal of Robotics Research}} \bibinfo{volume}{27}, \bibinfo{number}{6} (\bibinfo{year}{2008}), \bibinfo{pages}{737--757}.
\newblock


\bibitem[\protect\citeauthoryear{Narvekar, Peng, Leonetti, Sinapov, Taylor, and Stone}{Narvekar et~al\mbox{.}}{2020}]%
        {Narvekar20}
\bibfield{author}{\bibinfo{person}{S. Narvekar}, \bibinfo{person}{B. Peng}, \bibinfo{person}{M. Leonetti}, \bibinfo{person}{J. Sinapov}, \bibinfo{person}{M.~E. Taylor}, {and} \bibinfo{person}{P. Stone}.} \bibinfo{year}{2020}\natexlab{}.
\newblock \showarticletitle{Curriculum learning for reinforcement learning domains: A framework and survey}.
\newblock \bibinfo{journal}{\emph{Journal of Machine Learning Research}} \bibinfo{volume}{21}, \bibinfo{number}{181} (\bibinfo{year}{2020}), \bibinfo{pages}{1--50}.
\newblock


\bibitem[\protect\citeauthoryear{Narvekar}{Narvekar}{2021}]%
        {Narvekar21}
\bibfield{author}{\bibinfo{person}{S.~S. Narvekar}.} \bibinfo{year}{2021}\natexlab{}.
\newblock \emph{\bibinfo{title}{Curriculum learning in reinforcement learning}}.
\newblock \bibinfo{thesistype}{Ph.D. Dissertation}. \bibinfo{school}{The University of Texas at Austin}.
\newblock
\newblock
\shownote{Order Number: 29605085.}


\bibitem[\protect\citeauthoryear{Nkhumise, Basu, Prescott, and Gilra}{Nkhumise et~al\mbox{.}}{2025}]%
        {Nkhumise25}
\bibfield{author}{\bibinfo{person}{R.~M. Nkhumise}, \bibinfo{person}{D. Basu}, \bibinfo{person}{T.~J. Prescott}, {and} \bibinfo{person}{A. Gilra}.} \bibinfo{year}{2025}\natexlab{}.
\newblock \showarticletitle{Studying Exploration in RL: An Optimal Transport Analysis of Occupancy Measure Trajectories}.
\newblock \bibinfo{journal}{\emph{Transactions on Machine Learning Research (TMLR)}} (\bibinfo{year}{2025}).
\newblock


\bibitem[\protect\citeauthoryear{Obando-Ceron, Ara{\'u}jo, Courville, and Castro}{Obando-Ceron et~al\mbox{.}}{2024}]%
        {Obando24}
\bibfield{author}{\bibinfo{person}{J. Obando-Ceron}, \bibinfo{person}{J.~G.~M. Ara{\'u}jo}, \bibinfo{person}{A. Courville}, {and} \bibinfo{person}{P.~S. Castro}.} \bibinfo{year}{2024}\natexlab{}.
\newblock \showarticletitle{On the consistency of hyper-parameter selection in value-based deep reinforcement learning}. In \bibinfo{booktitle}{\emph{Reinforcement Learning Conference (RLC)}}. \bibinfo{publisher}{Reinforcement Learning Journal (RLJ)}.
\newblock


\bibitem[\protect\citeauthoryear{Oller, Glasmachers, and Cuccu}{Oller et~al\mbox{.}}{2020}]%
        {Oller20}
\bibfield{author}{\bibinfo{person}{D. Oller}, \bibinfo{person}{T. Glasmachers}, {and} \bibinfo{person}{G. Cuccu}.} \bibinfo{year}{2020}\natexlab{}.
\newblock \showarticletitle{Analyzing reinforcement learning benchmarks with random weight guessing}. In \bibinfo{booktitle}{\emph{Proceedings of the 19th International Conference on Autonomous Agents and MultiAgent Systems}}. \bibinfo{pages}{975–982}.
\newblock


\bibitem[\protect\citeauthoryear{Osband, Doron, Hessel, Aslanides, Sezener, Saraiva, McKinney, Lattimore, Szepesvari, Singh, van Roy, Sutton, Silver, and van Hassel}{Osband et~al\mbox{.}}{2019}]%
        {Osband19}
\bibfield{author}{\bibinfo{person}{I. Osband}, \bibinfo{person}{Y. Doron}, \bibinfo{person}{M. Hessel}, \bibinfo{person}{J. Aslanides}, \bibinfo{person}{E. Sezener}, \bibinfo{person}{A. Saraiva}, \bibinfo{person}{K. McKinney}, \bibinfo{person}{T. Lattimore}, \bibinfo{person}{C. Szepesvari}, \bibinfo{person}{S. Singh}, \bibinfo{person}{B. van Roy}, \bibinfo{person}{R. Sutton}, \bibinfo{person}{D. Silver}, {and} \bibinfo{person}{H. van Hassel}.} \bibinfo{year}{2019}\natexlab{}.
\newblock \showarticletitle{Behaviour suite for reinforcement learning}.
\newblock \bibinfo{journal}{\emph{arXiv preprint arXiv:1908.03568}} (\bibinfo{year}{2019}).
\newblock
\showeprint[arxiv]{1908.03568}


\bibitem[\protect\citeauthoryear{Pathak, Agrawal, Efros, and Darrell}{Pathak et~al\mbox{.}}{2017}]%
        {Pathak17}
\bibfield{author}{\bibinfo{person}{D. Pathak}, \bibinfo{person}{P. Agrawal}, \bibinfo{person}{A.~A. Efros}, {and} \bibinfo{person}{T. Darrell}.} \bibinfo{year}{2017}\natexlab{}.
\newblock \showarticletitle{Curiosity-driven exploration by self-supervised prediction}. In \bibinfo{booktitle}{\emph{International conference on machine learning}}. PMLR, \bibinfo{pages}{2778--2787}.
\newblock


\bibitem[\protect\citeauthoryear{Rajan, Diaz, Guttikonda, Ferreira, Biedenkapp, von Hartz, and Hutter}{Rajan et~al\mbox{.}}{2023}]%
        {Rajan23}
\bibfield{author}{\bibinfo{person}{R. Rajan}, \bibinfo{person}{J.~L.~B. Diaz}, \bibinfo{person}{S. Guttikonda}, \bibinfo{person}{F. Ferreira}, \bibinfo{person}{A. Biedenkapp}, \bibinfo{person}{J.~O. von Hartz}, {and} \bibinfo{person}{F. Hutter}.} \bibinfo{year}{2023}\natexlab{}.
\newblock \showarticletitle{MDP playground: An analysis and debug testbed for reinforcement learning}.
\newblock \bibinfo{journal}{\emph{Journal of Artificial Intelligence Research}}  \bibinfo{volume}{77} (\bibinfo{year}{2023}), \bibinfo{pages}{821--890}.
\newblock


\bibitem[\protect\citeauthoryear{Rocchetta, Bellani, Compare, Zio, and Patelli}{Rocchetta et~al\mbox{.}}{2019}]%
        {Rocchetta19}
\bibfield{author}{\bibinfo{person}{R. Rocchetta}, \bibinfo{person}{L. Bellani}, \bibinfo{person}{M. Compare}, \bibinfo{person}{E. Zio}, {and} \bibinfo{person}{E. Patelli}.} \bibinfo{year}{2019}\natexlab{}.
\newblock \showarticletitle{A reinforcement learning framework for optimal operation and maintenance of power grids}.
\newblock \bibinfo{journal}{\emph{Applied energy}}  \bibinfo{volume}{241} (\bibinfo{year}{2019}), \bibinfo{pages}{291--301}.
\newblock


\bibitem[\protect\citeauthoryear{Schmidhuber, Hochreiter, and Bengio}{Schmidhuber et~al\mbox{.}}{1999}]%
        {Schmidhuber99}
\bibfield{author}{\bibinfo{person}{J. Schmidhuber}, \bibinfo{person}{S. Hochreiter}, {and} \bibinfo{person}{Y. Bengio}.} \bibinfo{year}{1999}\natexlab{}.
\newblock \showarticletitle{Evaluating benchmark problems by random guessing}.
\newblock In \bibinfo{booktitle}{\emph{A Field Guide to Dynamical Recurrent Networks}}. \bibinfo{publisher}{Wiley}, \bibinfo{pages}{1329--–1338}.
\newblock


\bibitem[\protect\citeauthoryear{Sciavicco and Siciliano}{Sciavicco and Siciliano}{2012}]%
        {Sciavicco12}
\bibfield{author}{\bibinfo{person}{L. Sciavicco} {and} \bibinfo{person}{B. Siciliano}.} \bibinfo{year}{2012}\natexlab{}.
\newblock \bibinfo{booktitle}{\emph{Modelling and control of robot manipulators}}.
\newblock \bibinfo{publisher}{Springer}.
\newblock


\bibitem[\protect\citeauthoryear{Siciliano, Khatib, and Kr{\"o}ger}{Siciliano et~al\mbox{.}}{2008}]%
        {Siciliano08}
\bibfield{author}{\bibinfo{person}{B. Siciliano}, \bibinfo{person}{O. Khatib}, {and} \bibinfo{person}{T. Kr{\"o}ger}.} \bibinfo{year}{2008}\natexlab{}.
\newblock \bibinfo{booktitle}{\emph{Springer handbook of robotics}}. Vol.~\bibinfo{volume}{200}.
\newblock \bibinfo{publisher}{Springer}.
\newblock


\bibitem[\protect\citeauthoryear{Siciliano, Sciavicco, Villani, and Oriolo}{Siciliano et~al\mbox{.}}{2009}]%
        {Siciliano09}
\bibfield{author}{\bibinfo{person}{B. Siciliano}, \bibinfo{person}{L. Sciavicco}, \bibinfo{person}{L. Villani}, {and} \bibinfo{person}{G. Oriolo}.} \bibinfo{year}{2009}\natexlab{}.
\newblock \bibinfo{booktitle}{\emph{Robotics: modelling, planning and control}}.
\newblock \bibinfo{publisher}{Springer}.
\newblock


\bibitem[\protect\citeauthoryear{Silver, Schrittwieser, Simonyan, Antonoglou, Huang, Guez, Hubert, Baker, Lai, Bolton, Chen, Lillicrap, Hui, Sifre, van~den Driessche, Graepel, and Hassabis}{Silver et~al\mbox{.}}{2017}]%
        {Silver17}
\bibfield{author}{\bibinfo{person}{D. Silver}, \bibinfo{person}{J. Schrittwieser}, \bibinfo{person}{K. Simonyan}, \bibinfo{person}{I. Antonoglou}, \bibinfo{person}{A. Huang}, \bibinfo{person}{A. Guez}, \bibinfo{person}{T. Hubert}, \bibinfo{person}{L. Baker}, \bibinfo{person}{M. Lai}, \bibinfo{person}{A. Bolton}, \bibinfo{person}{Y. Chen}, \bibinfo{person}{T. Lillicrap}, \bibinfo{person}{F. Hui}, \bibinfo{person}{L. Sifre}, \bibinfo{person}{G. van~den Driessche}, \bibinfo{person}{T. Graepel}, {and} \bibinfo{person}{D. Hassabis}.} \bibinfo{year}{2017}\natexlab{}.
\newblock \showarticletitle{Mastering the game of go without human knowledge}.
\newblock \bibinfo{journal}{\emph{nature}} \bibinfo{volume}{550}, \bibinfo{number}{7676} (\bibinfo{year}{2017}), \bibinfo{pages}{354--359}.
\newblock


\bibitem[\protect\citeauthoryear{Spong, Hutchinson, and Vidyasagar}{Spong et~al\mbox{.}}{2006}]%
        {Spong06}
\bibfield{author}{\bibinfo{person}{M.~W. Spong}, \bibinfo{person}{S. Hutchinson}, {and} \bibinfo{person}{M. Vidyasagar}.} \bibinfo{year}{2006}\natexlab{}.
\newblock \bibinfo{booktitle}{\emph{Robot modeling and control}}. Vol.~\bibinfo{volume}{3}.
\newblock \bibinfo{publisher}{John Wiley \& Sons}.
\newblock


\bibitem[\protect\citeauthoryear{Stulp and Schaal}{Stulp and Schaal}{2011}]%
        {Stulp11}
\bibfield{author}{\bibinfo{person}{F. Stulp} {and} \bibinfo{person}{S. Schaal}.} \bibinfo{year}{2011}\natexlab{}.
\newblock \showarticletitle{Hierarchical reinforcement learning with movement primitives}. In \bibinfo{booktitle}{\emph{2011 11th IEEE-RAS International Conference on Humanoid Robots}}. IEEE, \bibinfo{pages}{231--238}.
\newblock


\bibitem[\protect\citeauthoryear{Sutton and Barto}{Sutton and Barto}{2018}]%
        {Sutton18}
\bibfield{author}{\bibinfo{person}{R.~S. Sutton} {and} \bibinfo{person}{A.~G. Barto}.} \bibinfo{year}{2018}\natexlab{}.
\newblock \bibinfo{booktitle}{\emph{Reinforcement Learning: An Introduction} (\bibinfo{edition}{2} ed.)}.
\newblock \bibinfo{publisher}{MIT Press}.
\newblock


\bibitem[\protect\citeauthoryear{Tassa, Doron, Muldal, Erez, Li, de~Las~Casas, Budden, Abdolmaleki, Merel, Lefrancqm, Lillicrap, and Reidmiller}{Tassa et~al\mbox{.}}{2018}]%
        {Tassa18}
\bibfield{author}{\bibinfo{person}{Y. Tassa}, \bibinfo{person}{Y. Doron}, \bibinfo{person}{A. Muldal}, \bibinfo{person}{T. Erez}, \bibinfo{person}{Y. Li}, \bibinfo{person}{D. de Las~Casas}, \bibinfo{person}{D. Budden}, \bibinfo{person}{A. Abdolmaleki}, \bibinfo{person}{J. Merel}, \bibinfo{person}{A. Lefrancqm}, \bibinfo{person}{T. Lillicrap}, {and} \bibinfo{person}{M. Reidmiller}.} \bibinfo{year}{2018}\natexlab{}.
\newblock \showarticletitle{Deepmind control suite}.
\newblock \bibinfo{journal}{\emph{arXiv preprint arXiv:1801.00690}} (\bibinfo{year}{2018}).
\newblock
\showeprint[arxiv]{1801.00690}


\bibitem[\protect\citeauthoryear{Thomas, Chien, Tamar, Ojea, and Abbeel}{Thomas et~al\mbox{.}}{2018}]%
        {Thomas18}
\bibfield{author}{\bibinfo{person}{G. Thomas}, \bibinfo{person}{M. Chien}, \bibinfo{person}{A. Tamar}, \bibinfo{person}{J.~A. Ojea}, {and} \bibinfo{person}{P. Abbeel}.} \bibinfo{year}{2018}\natexlab{}.
\newblock \showarticletitle{Learning robotic assembly from cad}. In \bibinfo{booktitle}{\emph{IEEE International Conference on Robotics and Automation (ICRA)}}. \bibinfo{pages}{3524--3531}.
\newblock


\bibitem[\protect\citeauthoryear{Vahrenkamp, Asfour, Metta, Sandini, and Dillmann}{Vahrenkamp et~al\mbox{.}}{2012}]%
        {Vahrenkamp12}
\bibfield{author}{\bibinfo{person}{N. Vahrenkamp}, \bibinfo{person}{T. Asfour}, \bibinfo{person}{G. Metta}, \bibinfo{person}{G. Sandini}, {and} \bibinfo{person}{R. Dillmann}.} \bibinfo{year}{2012}\natexlab{}.
\newblock \showarticletitle{Manipulability analysis}. In \bibinfo{booktitle}{\emph{12th IEEE-RAS international conference on humanoid robots (humanoids 2012)}}. IEEE, \bibinfo{pages}{568--573}.
\newblock


\bibitem[\protect\citeauthoryear{Wang, Liu, and Li}{Wang et~al\mbox{.}}{2020}]%
        {Wang20}
\bibfield{author}{\bibinfo{person}{J. Wang}, \bibinfo{person}{Y. Liu}, {and} \bibinfo{person}{B. Li}.} \bibinfo{year}{2020}\natexlab{}.
\newblock \showarticletitle{Reinforcement learning with perturbed rewards}. In \bibinfo{booktitle}{\emph{Proceedings of the AAAI conference on artificial intelligence}}, Vol.~\bibinfo{volume}{34}. \bibinfo{pages}{6202--6209}.
\newblock


\bibitem[\protect\citeauthoryear{West}{West}{2021}]%
        {West21}
\bibfield{author}{\bibinfo{person}{R.~M. West}.} \bibinfo{year}{2021}\natexlab{}.
\newblock \showarticletitle{Best practice in statistics: Use the Welch t-test when testing the difference between two groups}.
\newblock \bibinfo{journal}{\emph{Annals of clinical biochemistry}} \bibinfo{volume}{58}, \bibinfo{number}{4} (\bibinfo{year}{2021}), \bibinfo{pages}{267--269}.
\newblock


\bibitem[\protect\citeauthoryear{Whiteson, Tanner, Taylor, and Stone}{Whiteson et~al\mbox{.}}{2011}]%
        {Whiteson11}
\bibfield{author}{\bibinfo{person}{S. Whiteson}, \bibinfo{person}{B. Tanner}, \bibinfo{person}{M.~E. Taylor}, {and} \bibinfo{person}{P. Stone}.} \bibinfo{year}{2011}\natexlab{}.
\newblock \showarticletitle{Protecting against evaluation overfitting in empirical reinforcement learning}. In \bibinfo{booktitle}{\emph{2011 IEEE symposium on adaptive dynamic programming and reinforcement learning (ADPRL)}}. IEEE, \bibinfo{pages}{120--127}.
\newblock


\bibitem[\protect\citeauthoryear{Yang, Zhao, Li, and Zomaya}{Yang et~al\mbox{.}}{2020}]%
        {Yang20}
\bibfield{author}{\bibinfo{person}{T. Yang}, \bibinfo{person}{L. Zhao}, \bibinfo{person}{W. Li}, {and} \bibinfo{person}{A.~Y. Zomaya}.} \bibinfo{year}{2020}\natexlab{}.
\newblock \showarticletitle{Reinforcement learning in sustainable energy and electric systems: A survey}.
\newblock \bibinfo{journal}{\emph{Annual Reviews in Control}}  \bibinfo{volume}{49} (\bibinfo{year}{2020}), \bibinfo{pages}{145--163}.
\newblock


\bibitem[\protect\citeauthoryear{Yoshikawa}{Yoshikawa}{1985}]%
        {Yoshikawa85}
\bibfield{author}{\bibinfo{person}{T. Yoshikawa}.} \bibinfo{year}{1985}\natexlab{}.
\newblock \showarticletitle{Manipulability of robotic mechanisms}.
\newblock \bibinfo{journal}{\emph{The international journal of Robotics Research}} \bibinfo{volume}{4}, \bibinfo{number}{2} (\bibinfo{year}{1985}), \bibinfo{pages}{3--9}.
\newblock


\end{thebibliography}

%%%%%%%%%%%%%%%%%%%%%%%%%%%%%%%%%%%%%%%%%%%%%%%%%%%%%%%%%%%%%%%%%%%%%%%%

\newpage
\appendix
\clearpage
\onecolumn 
% \appendix
\section{Positional Errors at End-effector}\label{appx:ee_error}
This section demonstrates how joint errors propagate to the end effector. Given a planar serial chain with $n$ joints, the relationship between the end-effector velocities $\dot{x} \in \mathbb{R}^{2}$ and joint velocities $\dot{\theta}  \in \mathbb{R}^{n}$ is~\citep{Sciavicco12}:

\begin{equation}\label{forward_rel}
    \dot{x} = J(\theta)\dot{\theta}
\end{equation}

where $J(\theta) \in \mathbb{R}^{2 \times n}$ is the Jacobian matrix. The $k-$th column of $J(\theta)$ is given by:

\begin{equation}
\begin{aligned}
    J_{k}(\theta) &= \frac{\partial x}{\partial \theta_{k}} \\
    &= 
    \begin{bmatrix}
    -\sum_{i=k}^{n} l_i \sin\left( \sum_{j=1}^{i} \theta_{j} \right) \\
    \sum_{i=k}^{n} l_i \cos\left( \sum_{j=1}^{i} \theta_{j} \right)
    \end{bmatrix}
\end{aligned}
\end{equation}

whose Euclidean norm is

\begin{equation}\label{jacobian_column}
\begin{aligned}
    \left\| J_{k}(\theta)\right\|_{2} &= \sqrt{\left( \sum_{i=k}^{n} l_i \sin\left( \sum_{j=1}^{i} \theta_{j} \right) \right)^{2} +  \left( \sum_{i=k}^{n} l_i \cos\left( \sum_{j=1}^{i} \theta_{j} \right) \right)^{2}} \\
    &= \sqrt{\sum_{i=k}^{n} l_i^{2} + 2 \sum_{k\leq i < h \leq n}^{n}l_il_h\cos\left( \sum_{j=1}^{i} \theta_{j} - \sum_{j=1}^{h} \theta_{j}\right) } \\
    & \leq \sum_{i=k}^{n} l_i
\end{aligned}
\end{equation}

Given a small joint error, we can use Equation~\ref{forward_rel} to estimate the error at the end-effector as:

\begin{equation}\label{forward_rel_approx}
\begin{aligned}
    \delta{x} &\approx J(\theta) \delta{\theta} \\
    &=  \sum_{k=1}^{n} J_{k} \delta{\theta_k}
\end{aligned}
\end{equation}

Therefore, the magnitude of the $|| \delta{x} ||_{2}$ can be bounded using the triangular inequality as follows:

\begin{equation}\label{forward_rel_bounds}
\begin{aligned}
    \left\| \delta{x} \right\|_{2} &= \left\| \sum_{k=1}^{n} J_{k} \delta{\theta_k} \right\|_{2} \\
    & \leq \sum_{k=1}^{n} \left\| J_{k} \delta{\theta_k} \right\|_{2} \\
    & = \sum_{k=1}^{n} \left\| J_{k} \right\|_{2} | \delta{\theta_k}| \\
\end{aligned}
\end{equation}

If $| \delta{\theta_k}| \leq \epsilon$ for $k=1,\cdots,n$, and we substitute Equation~\ref{jacobian_column} into Equation~\ref{forward_rel_bounds}, then

\begin{equation}\label{forward_rel_bounds_all}
\begin{aligned}
    \left\| \delta{x} \right\|_{2} &\leq \sum_{k=1}^{n} \left\| J_{k} \right\|_{2} \epsilon \\
    &\leq \epsilon \sum_{k=1}^{n}\sum_{i=k}^{n} l_i \\
    & = \epsilon \sum_{i=1}^{n}\sum_{k=1}^{i} l_i \\
    &= \epsilon \sum_{i=1}^{n} il_i
\end{aligned}
\end{equation}

The final results of Equation~\ref{forward_rel_bounds_all} demonstrate that the error magnitude at the end-effector is exacerbated by both the number of joints $n$ and link lengths $l_i$. Therefore, the task complexity should increase when degree of freedom $n$ increase and/or when the link lengths $l_i$ increases, since an RL agent has to deal with increased error sensitivity. 

% \section{Learning}
\section{SAC Architecture}\label{appx:sac_arch}
In this section, we specify the architecture of the SAC algorithm used to produce the learning performances in Figure 2. Table~\ref{sac_con} entails a list of the configuration parameters for the algorithm. The ADAM~\citep{Kingma17} optimiser was used in both the actor and critic (neural-network) models.   
\begin{table}[h!]
\caption{SAC Hyperparameters.}\label{sac_con}
\centering
\begin{tabular}{l|cccr}
\hline
Parameter & Value \\
\hline
learning rate & $0.001$ \\
discount factor ($\gamma$) & $0.99$\\
temperature coefficient ($\alpha$) & $0.2$\\
number of hidden layers (all networks)  & 2 \\
number of hidden units per layer & 256 \\
number of samples per minibatch & 64 \\
nonlinearity & ReLU \\
target smoothing coefficient ($\tau$) & $0.005$ \\
Experience Replay size & $10^{6}$\\
% target update interval & 1 \\
% gradient steps & 1 \\
\hline
\end{tabular}
\end{table}

\section{Estimation Methods}\label{appx:stats_sig}
\subsection{PIC and POIC values}
This section describes how the mean and standard deviation of the PIC and POIC values presented in Table 1 were determined. We used bootstrap resampling of the $10^{4}$ samples to estimate the distribution of PIC and POIC. This allowed us in addition to compute the confidence intervals shown in Table~\ref{stats_pic_poic}.
\begin{table*}[h!]
\caption{Complete PIC and POIC values based on bootstrapping.}\label{stats_pic_poic}
\centering
\begin{tabular}{l|l|l|l|l}
\hline
Rewards & Tasks & Metric & Value & CI ($95\%$) \\
\hline
Dense & \textit{1-link} $(L=1m)$ & PIC & $4.005 \pm 0.0085$ & $(3.979;4.012)$\\
Dense & \textit{1-link} $(L=1m)$ & POIC & $(2.628\pm 0.056)\cdot10^{-3}$ & $(2.520;2.738)\cdot10^{-3}$\\
\hline
Dense & \textit{1-link} $(L=1.65m)$ & PIC & $4.153\pm 0.0084$ & $(4.126;4.159)$\\
Dense & \textit{1-link} $(L=1.65m)$ & POIC & $(4.105\pm 0.085)\cdot10^{-3}$ & $(3.944;4.279)\cdot10^{-3}$\\
\hline
Dense & \textit{2-link} & PIC & $4.205\pm 0.0061$ & $(4.195;4.219)$\\
Dense & \textit{2-link} & POIC & $(0.722\pm 0.011)\cdot10^{-3}$ & $(0.701;0.746)\cdot10^{-3}$\\
\hline
Sparse & \textit{1-link} $(L=1m)$ & PIC & $0.0851 \pm 0.0004$ & $(0.0843;0.0859)$\\
Sparse & \textit{1-link} $(L=1m)$ & POIC & $(1.982\pm 0.034)\cdot10^{-3}$ & $(1.911;2.047)\cdot10^{-3}$\\
\hline
Sparse & \textit{1-link} $(L=1.65m)$ & PIC & $0.0713\pm 0.0002$ & $(0.0708;0.0718)$\\
Sparse & \textit{1-link} $(L=1.65m)$ & POIC & $(1.197\pm 0.031)\cdot10^{-3}$ & $(1.106;1.233)\cdot10^{-3}$\\
\hline
Sparse & \textit{2-link} & PIC & $0.0460\pm 6.37\cdot10^{-5}$ & $(0.0458;0.0461)$\\
Sparse & \textit{2-link} & POIC & $(0.946\pm 0.0079)\cdot10^{-3}$ & $(0.932;0.962)\cdot10^{-3}$\\
\hline
\end{tabular}
\end{table*}

\subsection{Performance Distribution Plots}
\begin{figure*}[ht!]
\centering
\begin{tabular}{@{}c c@{}}
    \raisebox{1cm}{\rotatebox{90}{\textbf{1-link arm [$1.0$] (sparse)}}}  & %\hspace{0.2cm} & %3cm 
    \includegraphics[width=0.8\textwidth]{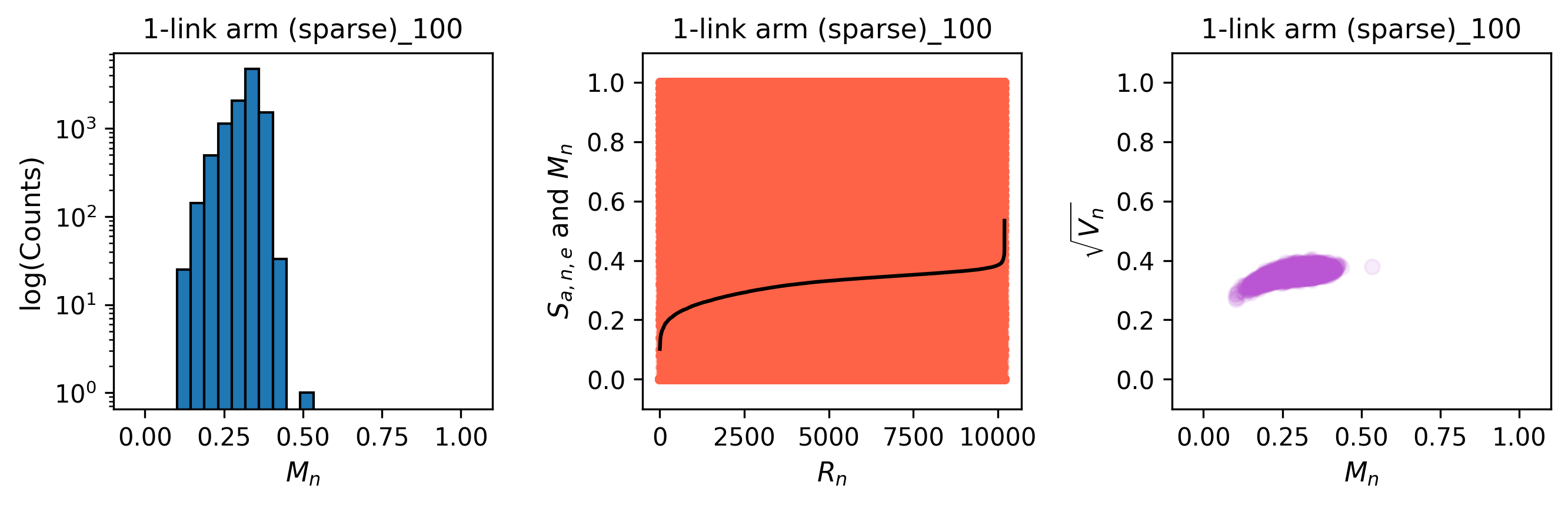} \\

    \raisebox{1cm}{\rotatebox{90}{\textbf{1-link arm [$1.65$] (sparse)}}}  & %\hspace{0.cm} & %3cm
    \includegraphics[width=0.8\textwidth]{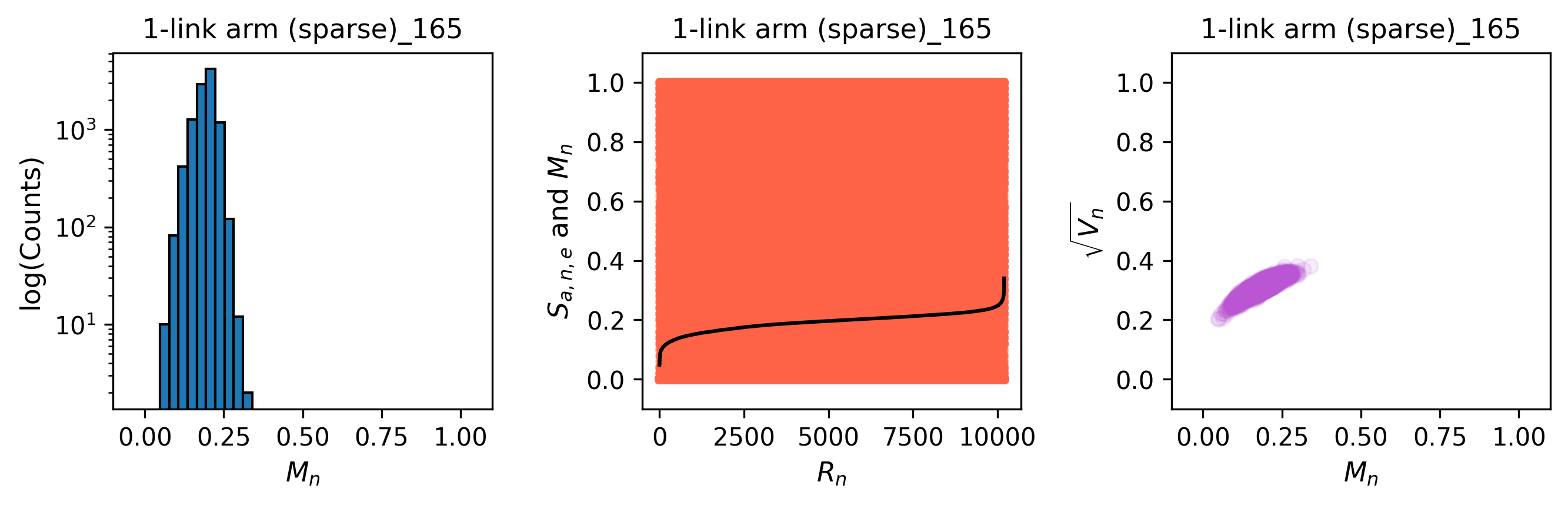} \\
\end{tabular}
\caption{Performance distribution plots for the tasks \textit{1-link} \textit{(L=1.0m)} and \textit{1-link} \textit{(L=1.65m)} arms with sparse rewards. The performance is normalised using min-max scaling to allow tasks to have the same range. Note that $a$ in $S_{a,n,e}$ represents the neural network architecture which is the same across all panels.}\label{dense_performance_plots}
\end{figure*}

This section presents performance distribution plots (in Figure~\ref{dense_performance_plots} ) for the tasks \textit{1-link} \textit{(L=1.0m)} and \textit{1-link} \textit{(L=1.65m)} arms with sparse rewards. 

\subsection{Statistical Significance Testing}
We depict results for using Welch's t-test~\citep{West21} for significance assessment. Table~\ref{appx:pic_vals} presents the results. The negative statistic denotes the incorrect order of task hardness according to PIC/POIC. We observe that the p-values are less than $0.005$, which makes the differences of PIC and POIC values across tasks statistically significant.
\begin{table*}[h!]
\caption{
Statistical significance between PIC and POIC values using Welch's t-test. Results are presented as: \textit{statistic (p-value)}. D stands for dense-rewards and S stands for sparse-rewards.}
\centering
\small
\begin{tabular}{| c | c | c | c | c | c |}
\hline
     & \textit{1-link [1.0](D)} & \textit{1-link [1.65](D)}  & \textit{2-link [0.95,1.7](D)} & \textit{1-link [1.0](S)} & \textit{1-link [1.65](S}) \\
\hline 
\multicolumn{6}{c}{\textbf{PIC}} \\
\hline
\textit{1-link [1.0](D)} & 0.0 (1.0) & - & - & - & - \\ %dense rewards
\hline
1\textit{-link [1.65](D)} & -20.1 (2.3$\times10^{-7}$) & 0.0 (1.0) & - & - & - \\
\hline
\textit{2-link [0.95,1.7](D)} & -29.3 (2.9$\times10^{-8}$) & -8.6 (2.6$\times10^{-5}$) & 0.0 (1.0) & - & - \\
\hline
\textit{1-link [1.0](S)} & 948 (4.2$\times10^{-12}$) & 646 (2.8$\times10^{-11}$) & 618 (3.4$\times10^{-11}$) & 0.0 (1.0) & - \\ %sparse rewards
\hline
\textit{1-link [1.65](S)} & 955 (5.7$\times10^{-12}$) & 649 (3.1$\times10^{-11}$) & 621 (3.7$\times10^{-11}$) & 25.4 (4.7$\times10^{-8}$) & 0.0 (1.0) \\
\hline
\textit{2-link [0.95,1.7](S)} & 964 (6.9$\times10^{-12}$) & 654 (3.3$\times10^{-11}$) & 626 (3.9$\times10^{-11}$) & 85.7 (6.5$\times10^{-8}$) & 84.5 (3.3$\times10^{-8}$) \\    
\hline
\multicolumn{6}{c}{\textbf{POIC}} \\
\hline
\textit{1-link [1.0](D)} & 0.0 (1.0) & - & - & - & - \\ %dense rewards
\hline
\textit{1-link [1.65](D)} & -17.7 (1.2$\times10^{-7}$) & 0.0 (1.0) & - & - & - \\
\hline
\textit{2-link [0.95,1.7](D)} & 30.8 (4.5$\times10^{-6}$) & 59.7 (2.6$\times10^{-7}$) & 0.0 (1.0) & - & - \\
\hline
\textit{1-link [1.0](S)} & 8.1 (4.8$\times10^{-5}$) & 27.9 (3.3$\times10^{-9}$) & -24.2 (3.4$\times10^{-5}$) & 0.0 (1.0) & - \\ %sparse rewards
\hline
\textit{1-link [1.65](S)} & 22.3 (4.6$\times10^{-6}$) & 49.0 (6.6$\times10^{-8}$) & -21.6 (1.5$\times10^{-6}$) & 14.4 (2.3$\times10^{-5}$) & 0.0 (1.0) \\
\hline
\textit{2-link [0.95,1.7](S)} & 27.2 (7.5$\times10^{-6}$) & 55.8 (3.4$\times10^{-7}$) & -17.9 (9.9$\times10^{-8}$) & 19.9 (2.4$\times10^{-5}$) & 11.3 (5.0$\times10^{-5}$) \\ 

\hline

\end{tabular}
\label{appx:pic_vals}
\end{table*}

\subsection{Initialisation Methods}\label{initialisation_methods}
\begin{figure*}[h!]
\centering
\begin{tabular}{@{}c @{}c c@{}}
    \includegraphics[width=0.29\textwidth]{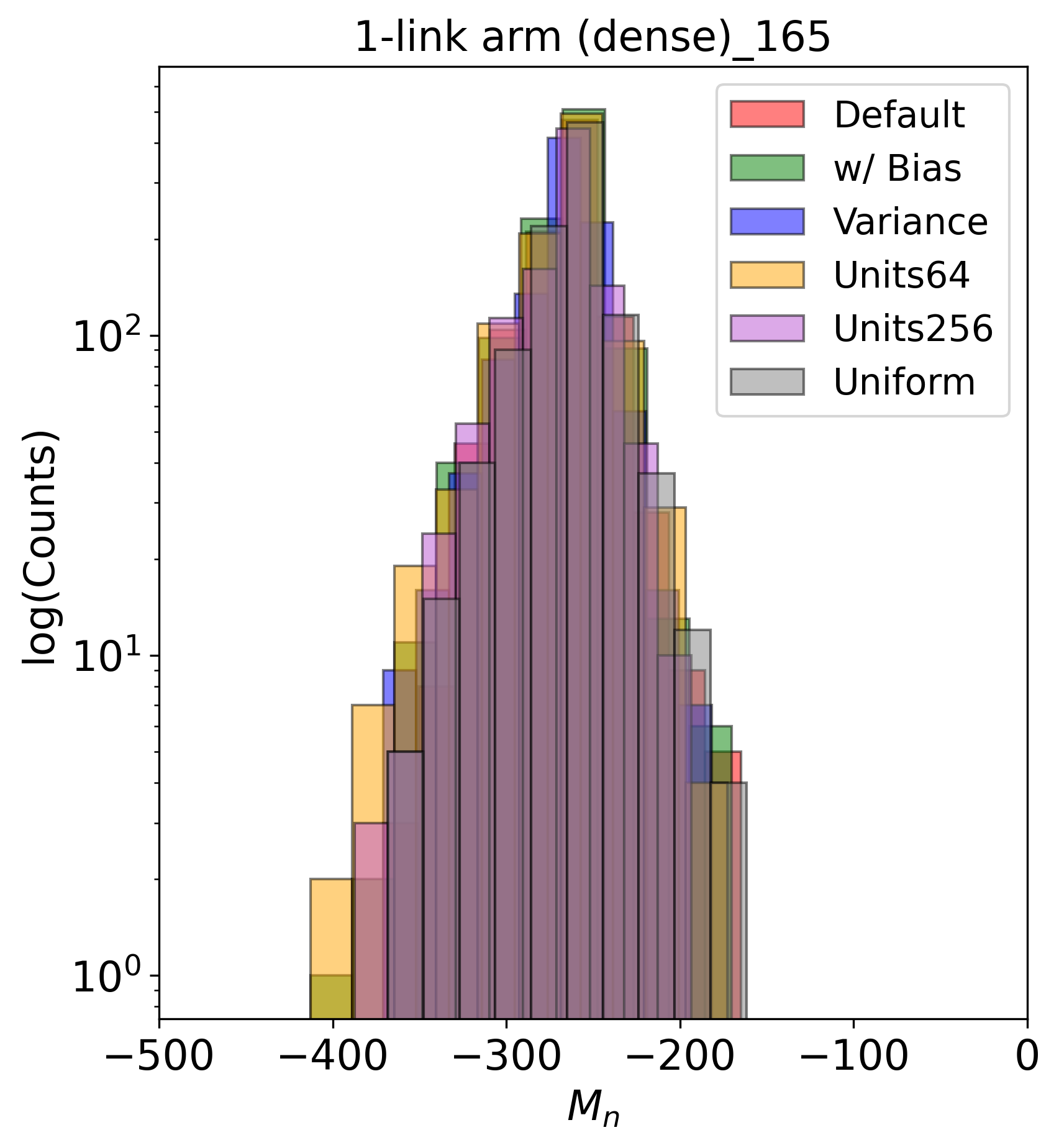} &
    \includegraphics[width=0.3\textwidth]{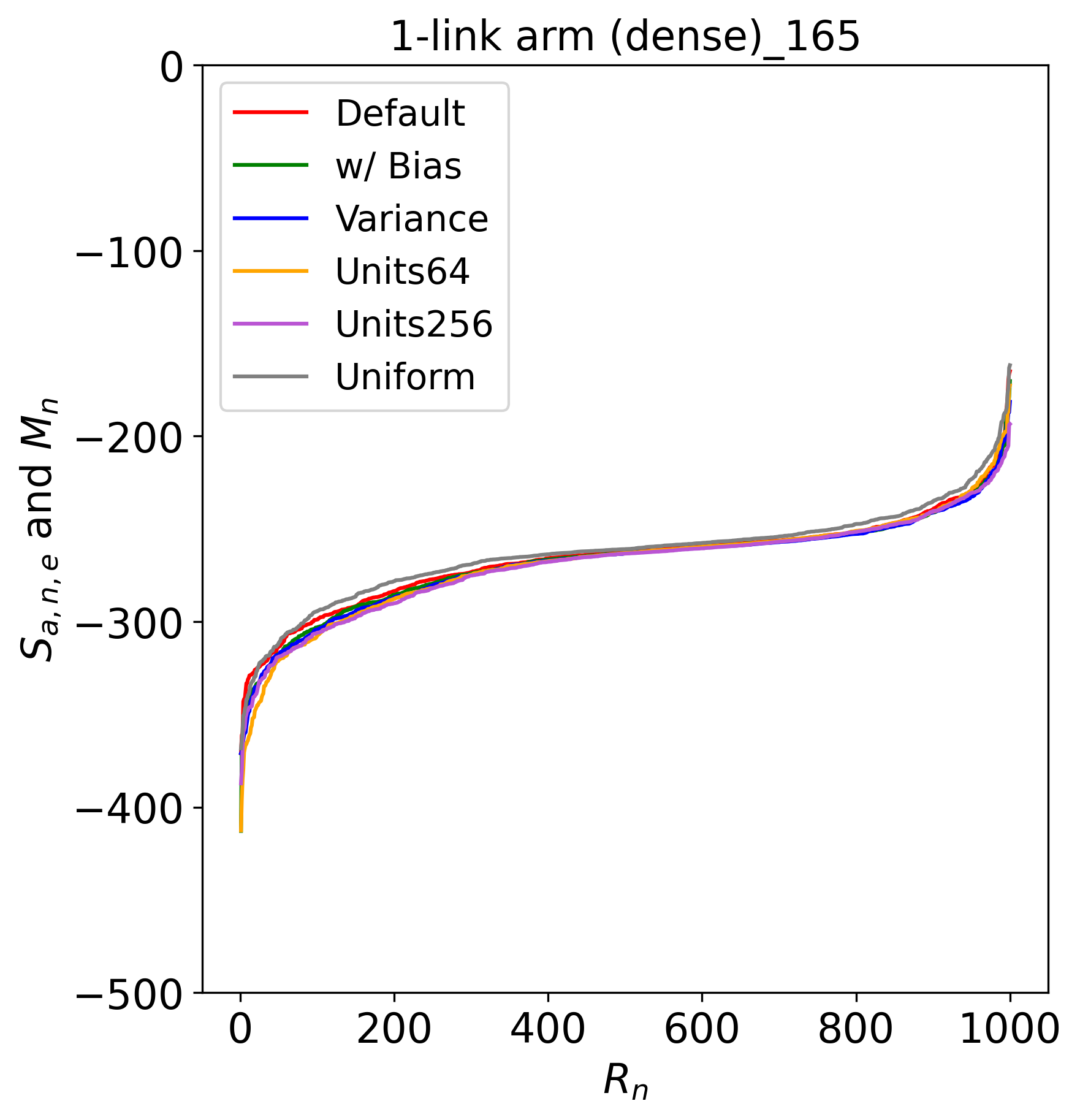} 
    &
    \includegraphics[width=0.29\textwidth]{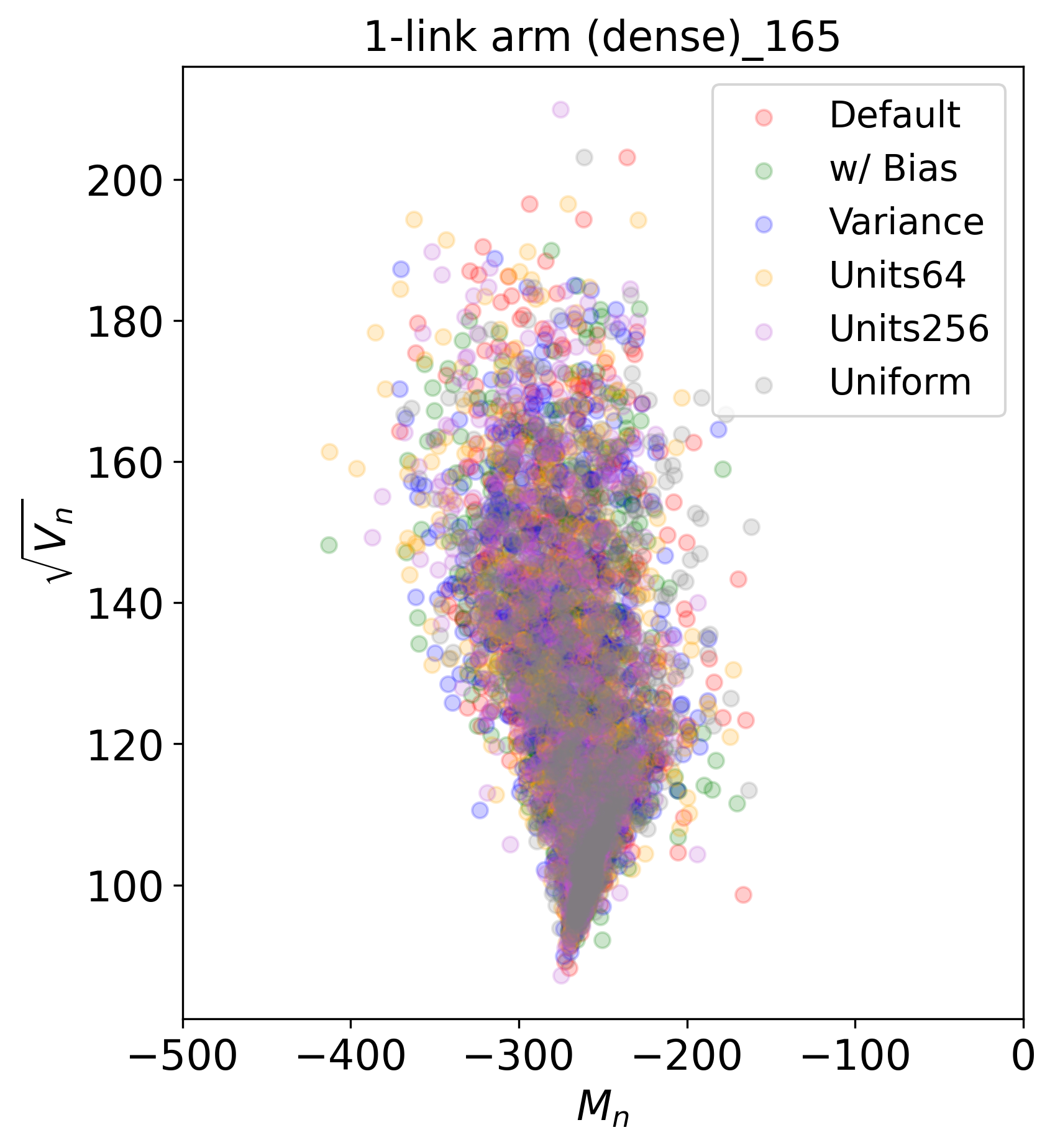} 
\end{tabular}
\caption{Performance distribution plots for the \textit{1-link} ($L=1.65$ m) arm with dense rewards across various policy network architectures and initialisation methods of weights. This shows how the random policies behave similarly across various settings. Note that $a$ in $S_{a,n,e}$ represents the neural network architecture which is the same across all panels.}\label{localisation_of_policies}
\end{figure*}

In RWG, we mainly used multivariate normal distribution prior $p(\theta) = \mathcal{N}(0,I)$, where $I \in \mathbb{R}^{d \times d}$ given weight vectors $\theta_n \in \mathbb{R}^{d}$, to sample weights $\theta_n \sim p(\theta)$. This was combined with a neural network architecture (NN) of 2 hidden layers (HL), each with 32 units (HU) and the NN is without bias, i.e. NN = [2HL, 32HU, w/o bias]. In Figure~\ref{localisation_of_policies}, we show that performance distributions for the tasks are unchanging when prior distribution is altered. We tested with the following cases:
\begin{enumerate}
    \item Default: $p(\theta) = \mathcal{N}(0,I)$ + NN = [2HL, 32HU, w/o bias]
    \item w/ Bias: $p(\theta) = \mathcal{N}(0,I)$ + NN = [2HL, 32HU, w/ bias]
    \item Variance: $p(\theta) = \mathcal{N}(0,2I)$ + NN = [2HL, 32HU, w/o bias]
    \item Units64: $p(\theta) = \mathcal{N}(0,I)$ + NN = [2HL, 64HU, w/o bias]
    \item Units256: $p(\theta) = \mathcal{N}(0,I)$ + NN = [2HL, 256HU, w/o bias]
    \item Uniform: $p(\theta) = Unif(-1,1)$ + NN = [2HL, 32HU, w/o bias]
    % \item $p(\theta) = \mathcal{N}(0,I)$ + NN = [2HL, 64HU, w/ bias]
\end{enumerate}
To attain the results in Figure~\ref{localisation_of_policies}, we used $N=10^3$ samples (i.e. random policies) in each setting, where each sample was deploy across $500$ episodes. 

\begin{figure}[h!]
\centering
    \includegraphics[width=0.35\textwidth]{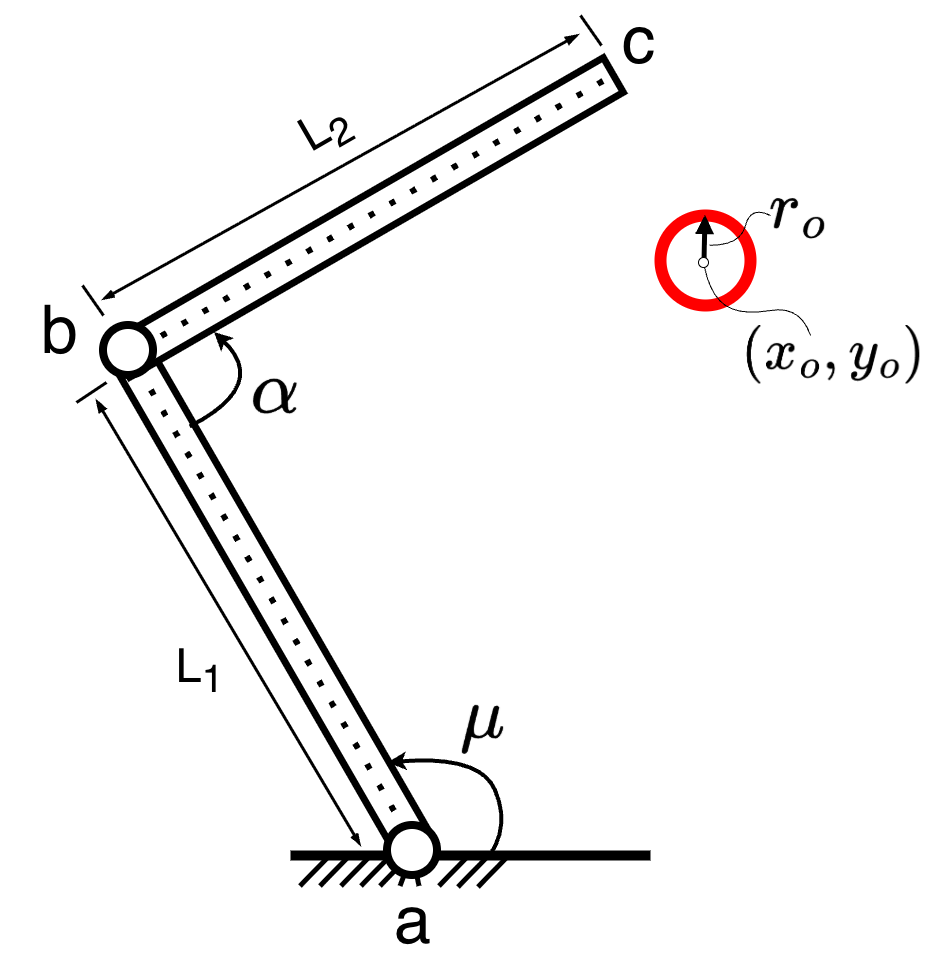} 
\caption{\textit{2-link} arm with obstacle in the workspace.}\label{arm_illustration_obstacle}
\end{figure}

{\section{2-link  arm with obstacle}}\label{2_link_with_obstacle}
We added another task setting, where an obstacle is introduced in the workspace of the \textit{2-link} arm with dense rewards. This enriches our framework with another structurally similar task. We expect that the task with an obstacle should be harder than without the obstacle. Figure~\ref{arm_illustration_obstacle} depicts the task setting. The red circle is the obstacle with a radius $r_o$ located at position $(x_o,y_o)$. Similar to other tasks, friction and gravity are ignored, and we assume all links have a thickness of $t_o$. The aim of the task is for the end-effector (arm-tip) to reach an arbitrary target (within the workspace) while avoiding collision with the obstacle. 

We can define the distance between the each link and the obstacle as: 
\begin{equation}\label{distance_between_link_obstacle}
\begin{aligned}
    d_{i} &= \min_{p \in link_i} \text{dist}(p, \mathcal{O}) \\
    &= \min_{p \in link_i} \left[ \lVert p - (x_o,y_o) \rVert - (r_o + 0.5t_o) \right]
\end{aligned}
\end{equation}
where $\mathcal{O}$ is the obstacle region defined by radius $r_o$ and $p$ is any point lying along the $i^{th}$ link. Starting from the ground (a) to the end-effector (c), the first link has $p = a + \frac{k}{N}(b-a)$, while the second link has $p = b + \frac{k}{N}(c-b)$. Note that $a,b,c$ represent coordinates of each joint and the end-effector. $N$ are the number of sample points along a link with index $k \in [0,N]$. Our definition of $p$ follows a uniform discretisation of a line segment. When $d_i \le 0$ (we set $d_i$ to zero if negative), the link and obstacle are in collision, while for $d_i > 0$ the link is collision free. 
% \begin{equation}\label{about_d}
% d_i = \begin{dcases}
%         0, &~\text{link and obstacle are in collision} \\
%         > 0, &~\text{collision free}
% \end{dcases}
% \end{equation}
In our setting, $N = 50$, $r_o = 0.02$ m, $t_o = 0.05$ m, $L_1 = 0.95$ m and $L_2 = 0.7$ m. 

The reward function that best captures the task aim is given by:
\begin{equation}\label{reward_function_dense_obstacle}
    r = -\omega_1\lVert P_{ee} - P_{g} \rVert_{2}^{2} - \omega_2\lVert action \rVert_{2}^{2} - \beta_1 \mathbb{I}_{collision} -  \beta_2 \sum_{i=1}^2 e^{-\alpha d_i}
\end{equation}
where $[\omega_1,\omega_2] = [1,1]$ are distance and control weights. $P_{ee}$ and $P_{g}$ are respectively end-effector and target/goal positions. $[\beta_1,\beta_2] = [10^3,5]$ are collision and proximity weights and $\alpha = 2.5$ is a smoothing factor. The collision penalty is 
\begin{equation}\label{collision_penalty}
\mathbb{I}_{collision} = \begin{dcases}
        1, &~\text{if any link collides with obstacle} \\
        0, &~\text{otherwise}
\end{dcases}
\end{equation}
while the proximity penalty is expressed as the sum of exponentials $\sum_{i=1}^2 e^{-\alpha d_i}$. During implementation, $d_i$ is set to $0$ if negative, so that $e^{-\alpha d_i} \in [0,1]$. This means $e^{-\alpha d_i}$ approaches unity when $d_i \to 0$ and approaches zero when $d_i \gg 0$. Note that the obstacle position was constant for the entire training run and the environment episodes terminated under the following three conditions:
\begin{enumerate}
    \item When collision occurs, i.e. $d_i = 0$. 
    \item When the end-effector is within threshold distance from the target, i.e. $\lVert P_{ee} - P_{g} \rVert_{2} \le 0.05$ m.
    \item When a maximum of $50$ steps is reached.
\end{enumerate}

\begin{figure}[th!]
\centering
    \includegraphics[width=0.45\textwidth]{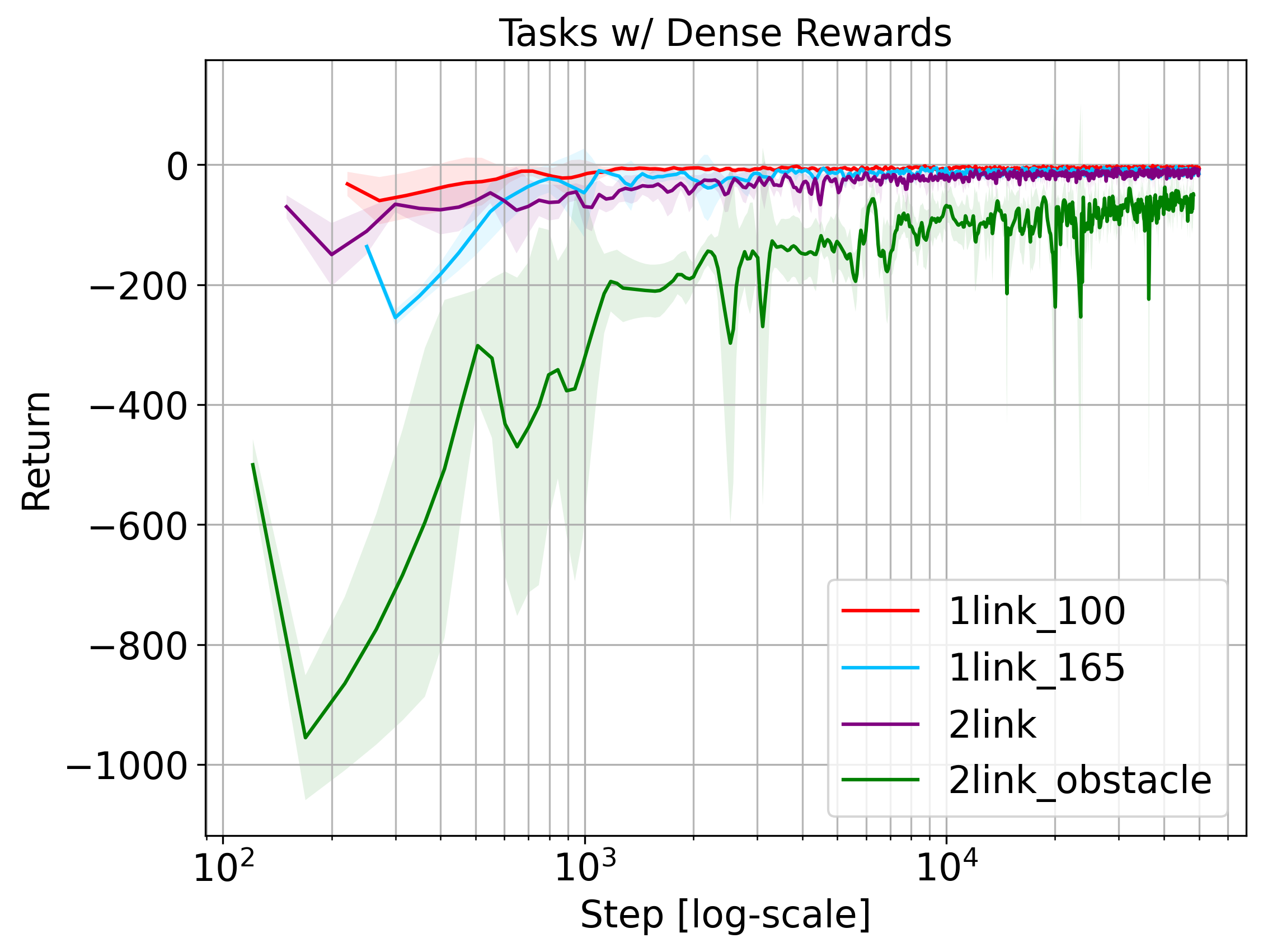}
\caption{Learning curves of SAC algorithm across tasks with dense rewards. To accommodate wide and varying ranges of steps, we draw the x-axis on a logarithmic scale to enhance interpretability. The results are obtained via evaluation of each task over 5 runs.
% Note that the plots seem to have different starting steps due to varying sampling rates. 
}\label{RL_learning_across_obstacle}
\end{figure}

The learning curves for this task along with others are presented in Figure~\ref{RL_learning_across_obstacle}. We can see from the plots that SAC takes longer to converge and converges at return values below the rest, especially \textit{2-link} arm with dense-rewards and no obstacle. This shows that the task is harder than its counterpart. 

\begin{figure*}[h!] 
\centering
     \raisebox{1.0cm}{\rotatebox{90}{\textbf{2-link arm (dense)}}}
     \hspace{.125cm}
     \begin{subfigure}[b]{0.8\textwidth}
         \centering
         \includegraphics[width=\textwidth]{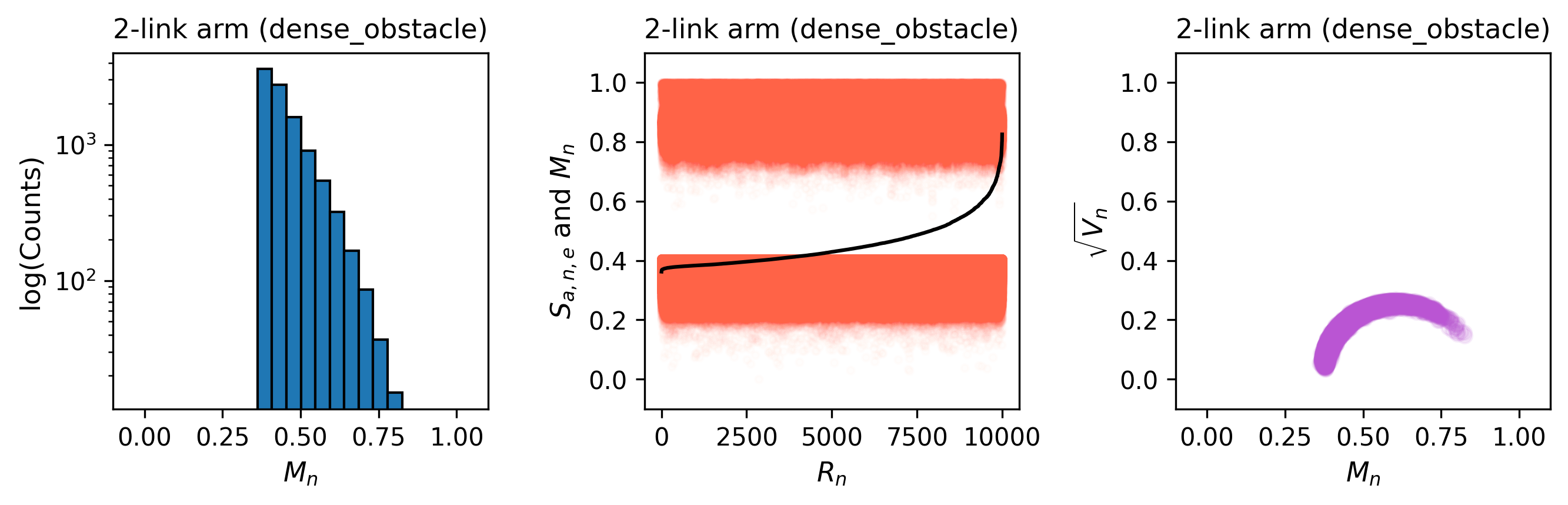} 
         \caption{}
     \end{subfigure}
\caption{Performance distribution plots for the~\textit{2-link} arm with dense rewards and an obstacle in the workspace. The left column shows a histogram of mean performances of the random policies (\textit{Log-scale histogram of} $M_{n}$). The middle column depicts \textit{mean performance curves} in black, i.e. mean performance $M_{n}$ vs rank $R_{n}$. Moreover, all the cumulative rewards of the policies $s_{a,n,e}$ across the trials are represented by red dots (behind the black curve), where $a$ represents the neural network architecture (same across all panels), $n$ and $e$ are as explained in the main text. The right column displays plots of standard deviation $\sqrt{V_{n}}$ vs mean performance $M_{n}$ (often referred to as \textit{variance distribution}). The plots were made using $10^{4}$ random policies. }
\label{performance_plots_obstacle}
\end{figure*}

In Figure~\ref{performance_plots_obstacle}, the \textit{mean performance histogram} is right-skewed with the mode (i.e. most frequent value) being less than $50\%$. 
% This shows that most policies attained less than $50\%$ mean performance. 
The \textit{mean performance curve} plot depicts two bands on episodic returns, one below $0.4$ and another above $0.6$. When collision with the obstacle did not occur, policies managed to reach peak performance in some episodes thus behaving similar to a \textit{2-link} arm (dense-rewards) without obstacles. However, when collision occurred, performance was limited to $40\%$ of the maximum return. This underscores the effect of an obstacle in the environment. The \textit{variance distribution} illustrates how the highest variance was experienced at approximately the centre of the range of the mean performance, and the lowest variance was below the centre. This shows that when the obstacle was encountered ($M_n < 0.5$), performance was bounded in comparison to collision-free episodes. Figure~\ref{performance_plots_obstacle} is consistent with Figure~\ref{RL_learning_across_obstacle} in showing that this task is harder than the \textit{2-link} arm with dense-rewards and without any obstacles. 
\vspace{1em}

By studying Table~\ref{pic_vals_obstacle}, we notice that the PIC value of the \textit{2-link} arm task with an obstacle is the lowest amongst tasks in dense-rewards settings. This meets our expectations, since the metric implies that the task is the hardest amongst others with dense rewards. However, the POIC value of the task is the highest amongst its counterparts. This denotes that finding the optimal behaviour in this task is easier than all other tasks. This conflicts with results in Figure~\ref{RL_learning_across_obstacle}, thus showing another instance of unreliable task hardness ranking by the metric. Results of statistical significance using the Welch's t-test~\citep{Delacre17} are presented in Table~\ref{appx:pic_vals_obstacle}. This shows strong statistical significance since the p-values of the t-statistic are in the orders of $10^{-5}$ below a typical cut-off p-value of $0.005$~\citep{Benjamin18}.
\begin{table}[h!]
\caption{
PIC and POIC values with $N = 10^{4}$ samples (random policies). High PIC and POIC values correspond to easier tasks, while low values correspond to harder tasks.}
\centering
\begin{tabular}{| c | c | c | c |}
\hline
Rewards & Arm [dim] & PIC $\left(\times 10^{-3}\right)$ & POIC $\left(\times 10^{-3}\right)$ \\
\hline \hline
\multirow{4}{*}{Dense} & 1-link [1.0] & $4005\pm8.5$ & $2.628\pm0.056$ \\ %dense rewards
    & 1-link [1.65] & $4153\pm8.4$ & $4.105\pm0.085$ \\
    & 2-link [0.95,1.7] & $4200\pm6.1$ & $0.725\pm0.011$ \\
    & 2-link [obstacle] & $3671\pm14.3$ & $21.31\pm0.33$ \\
\hline
\multirow{3}{*}{Sparse} & 1-link [1.0] & $85.11\pm0.4$ & $1.958\pm0.034$ \\ %sparse rewards
    & 1-link [1.65] & $71.21\pm0.2$ & $1.197\pm0.031$ \\
    & 2-link [0.95,1.7] & $45.95\pm0.0$ & $0.946\pm0.0079$ \\      
\hline
\end{tabular}
\label{pic_vals_obstacle}
\end{table}

\begin{table*}[h!]
\caption{
Statistical significance between PIC and POIC values using Welch's t-test. Results are presented as: \textit{statistic (p-value)}. These are for only tasks with dense-rewards.}
\centering
\small
\begin{tabular}{| c | c | c | c | c |}
\hline
     & \textit{1-link [1.0]} & \textit{1-link [1.65]}  & \textit{2-link [0.95,1.7]} & \textit{2-link [obstacle]}  \\
\hline 
\multicolumn{5}{c}{\textbf{PIC}} \\
\hline
\textit{1-link [1.0]} & 0.0 (1.0) & - & - & - \\ %dense rewards
\hline
1\textit{-link [1.65]} & -20.1 (2.3$\times10^{-7}$) & 0.0 (1.0) & - & - \\
\hline
\textit{2-link [0.95,1.7]} & -29.3 (2.9$\times10^{-8}$) & -8.6 (2.6$\times10^{-5}$) & 0.0 (1.0) & -  \\
\hline
\textit{2-link [obstacle]} & 37.4 (5.4$\times10^{-9}$) & 48.5 (3.8$\times10^{-11}$) & 55.4 (1.3$\times10^{-11}$) & 0.0 (1.0)  \\ %sparse rewards
\hline
   
\multicolumn{5}{c}{\textbf{POIC}} \\
\hline
\textit{1-link [1.0]} & 0.0 (1.0) & - & - & - \\ %dense rewards
\hline
\textit{1-link [1.65]} & -17.7 (1.2$\times10^{-7}$) & 0.0 (1.0) & - & - \\
\hline
\textit{2-link [0.95,1.7]} & 30.8 (4.5$\times10^{-6}$) & 59.7 (2.6$\times10^{-7}$) & 0.0 (1.0) & - \\
\hline
\textit{2-link [obstacle]} & -85.1 (1.3$\times10^{-8}$) & 79.0 (2.6$\times10^{-8}$) & -97.5 (6.3$\times10^{-8}$) & 0.0 (1.0) \\ %sparse rewards
\hline

\end{tabular}
\label{appx:pic_vals_obstacle}
\end{table*}

% the reward function is
% \begin{equation}\label{reward_function_sparse}
%     r = \begin{dcases}
%         ~~~~0, &~\text{if } ~\lVert P_{ee} - P_{g} \rVert_{2} < 0.05 ~\text{meters} \\
%         -1, &~\text{otherwise}
%     \end{dcases}
% \end{equation}
% where $0.05$ meters is the threshold distance between the end-effector and target position. 

\newpage
\clearpage
\end{document}